\documentclass[10pt,a4paper,logo,twocolumn]{qwenapplication}
\usepackage[numbers]{natbib}
\usepackage{xurl}
\usepackage{graphicx}
\usepackage{subcaption}

\usepackage{booktabs} 
\usepackage{amsmath}
\usepackage{amssymb}
\usepackage{mathtools}
\usepackage{amsthm}
\usepackage{algorithm}
\usepackage{algorithmic}

\usepackage[capitalize,noabbrev]{cleveref}

\theoremstyle{plain}

\theoremstyle{definition}

\theoremstyle{remark}

\usepackage{bbm}
\usepackage{multirow}
\usepackage{pifont}

\title{COPE: Continual Personalization of LLMs under Sparse User Feedback via User Embeddings and Self-Evaluation}

\author[1,2]{Ruike Cao}
\author[2,*]{Fugen Yao}
\author[2]{Liang Dong}
\author[2]{Jian Xu}
\author[2]{Guanjun Jiang}
\author[1]{Li Xiao}

\affil[1]{University of Science and Technology of China}
\affil[2]{Qwen Business Unit of Alibaba}
\authornote{\textsuperscript{*}Corresponding author.}

\begin{abstract}
While Large Language Models (LLMs) have achieved remarkable results across various benchmarks, their alignment with normative values often results in homogenized responses that fail to address diverse user preferences.
Existing training-free methods often occupy valuable context windows through prompt engineering, while training-based methods typically remain static post-training, failing to support the continual optimization required in real-world settings.
To address these challenges, we propose \textbf{COPE} (\textbf{C}ontinual \textbf{O}ptimization with \textbf{P}ersonalized embedding and self-\textbf{E}valuation), a novel optimization framework tailored for real-world-motivated interaction settings with sparse user feedback.
Our framework assigns learnable personalized embeddings to each user and synergistically integrates preference capture, self-evaluation calibration, and personalized response optimization within a single update step.
A key innovation of our method is the use of self-evaluation to generate proxy rewards, enabling continuous model updates even when explicit user feedback is unavailable. Experiments show that COPE consistently outperforms strong training-free and training-based baselines under sparse feedback, and remains complementary to Retrieval-Augmented Prompting (RAP). Further analyses confirm COPE's reliable self-evaluation, meaningful preference patterns, stable general capabilities, and robustness under shifting preferences and alternative evaluators.\footnote{Code, data, and prompts are available at \url{https://github.com/Quark-Medical/COPE}.}
\end{abstract}

\begin{document}
\raggedbottom
\maketitle

\begingroup
\renewcommand{\thefootnote}{*}
\footnotetext[0]{\raggedright Corresponding author: Fugen Yao
(\href{mailto:fugen.yfg@alibaba-inc.com}{\nolinkurl{fugen.yfg@alibaba-inc.com}}).}
\endgroup

\section{Introduction}
In recent years, Large Language Models (LLMs) have achieved rapid advances across challenging benchmarks~\citep{yang2025qwen3, team2025kimi, liu2025deepseek} and are increasingly used as everyday assistants~\citep{mcdaniel2025artificial}. Beyond general capabilities in knowledge answering and reasoning, user adoption also depends on whether responses reflect individual preferences~\citep{guan2025survey}. However, existing LLMs are commonly aligned through Reinforcement Learning from Human Feedback (RLHF)~\cite{ouyang2022training} on carefully curated normative preferences, implicitly assuming a shared value system across users~\citep{bai2022constitutional, siththaranjan2023distributional}. As a result, they tend to produce homogenized responses that reflect the ``average'' of group preferences and may underrepresent minority preferences~\citep{poddar2024personalizing, chen2025pal}.
To address this limitation, prior work has explored both training-free and training-based personalization. Training-free methods mainly rely on in-context signals, such as retrieved dialogue histories in Retrieval-Augmented Prompting (RAP)~\citep{salemi2024lamp, tavakoli2025beyond}, summarized user profiles in Profile-Augmented Prompting (PAP)~\citep{liu2024once, qiu2025measuring}, or user-specified preferences~\citep{lee2024aligning}. Although these methods avoid model updates, they consume valuable context space and may suffer from retrieval noise or information loss during summarization~\citep{liu2025survey}. Training-based approaches learn personalization through optimization, including Multi-Objective Reinforcement Learning (MORL) over predefined preference dimensions~\citep{wu2023fine, zhou2024beyond}, user-specific PEFT modules~\citep{zhang2024personalized, liu2025exploring}, and reinforcement learning based interaction modeling such as RLPA~\citep{zhao2025teaching}. However, MORL and RLPA rely on predefined preference dimensions or discrete user types, PEFT-style methods face data and scaling constraints, and most remain static after offline training, limiting adaptation to evolving user preferences~\citep{liu2025survey}.

To overcome these limitations, we consider a real-world-motivated personalized interaction setting in which a diverse population of users queries an assistant and expects responses aligned with their individual preferences, while providing explicit feedback only sparsely. We propose \textbf{COPE} (\textbf{C}ontinual \textbf{O}ptimization with \textbf{P}ersonalized embedding and self-\textbf{E}valuation), which assigns each user a learnable personalized embedding to guide user-specific response generation. COPE optimizes the assistant through a unified update step that integrates three complementary objectives: \textbf{\emph{Supervised Fine-Tuning for Preference Capture}}, which uses available feedback to encode user preferences into personalized embeddings; \textbf{\emph{Reinforcement Learning for Self-Evaluation Calibration}}, which aligns the model's own feedback prediction with real user feedback and learns proxy rewards; and \textbf{\emph{Reinforcement Learning for Personalized Response Optimization}}, which improves response generation using real feedback when available and proxy rewards otherwise. Executed as the optimization phase of the \emph{``Interact-Collect-Optimize''} loop, this update enables continuous model optimization under sparse feedback. Appendix~\ref{app:related_work} further discusses how COPE relates to and differs from recent user-embedding, prefix-based, and feedback-driven personalization methods~\citep{ning2025user,huber2025embedding,kim2026spring,ma2026personalizing}. Our contributions are summarized as follows:
\begin{enumerate}
    \item We establish a real-world-motivated setting for continual personalization under sparse user feedback and instantiate it on PersonaLens with a controlled chronological traversal protocol, bridging static personalization benchmarks with controlled evaluation of continual model optimization.
    \item We propose COPE, a continual optimization framework that maintains learnable personalized embeddings as compact user-specific preference representations and updates these embeddings together with the assistant model in the \emph{``Interact-Collect-Optimize''} loop. Calibrated self-evaluation supplies proxy rewards for interactions without explicit feedback, allowing the optimization loop to keep using collected data when real user feedback is sparse.
    \item Experiments show that COPE outperforms several strong training-free and training-based baselines across feedback probabilities and remains complementary to retrieval-based prompting. Analyses further show that the self-evaluation aligns with user feedback, the learned embeddings capture meaningful preference patterns, the optimized model largely preserves general capabilities and remains effective under preference shifts, and the gains hold under alternative evaluators.
\end{enumerate}
\begin{figure*}[ht]
    \centering
    \includegraphics[width=1\linewidth]{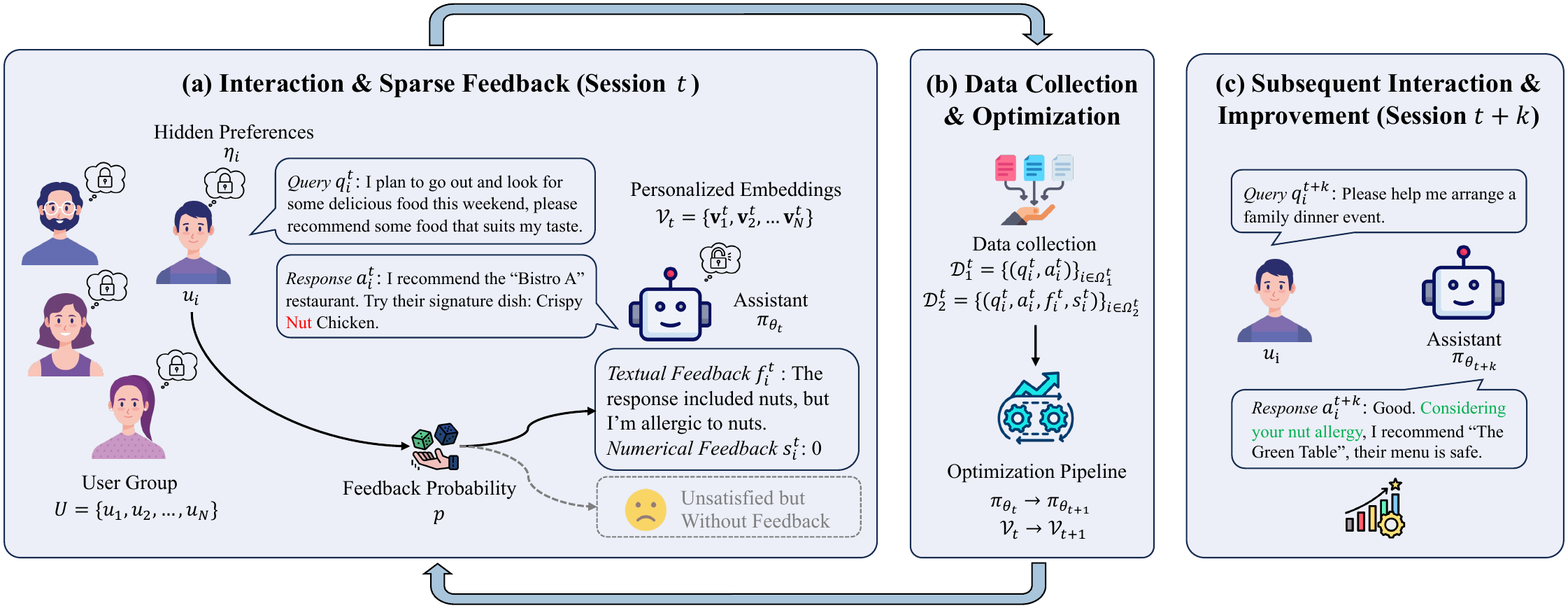}
    \caption{Illustration of the dynamic interaction scenario and the iterative \textit{``Interact-Collect-Optimize''} loop. (a) \textbf{Interaction \& Sparse Feedback:} During session $t$, for each user, the assistant $\pi_{\theta_t}$ generates a response to the query, guided by the user-specific personalized embedding. If it fails to align with the user's hidden preferences $\eta_i$, explicit feedback (textual $f_i^t$ and numerical $s_i^t$) is provided sparsely with probability $p$. (b) \textbf{Data Collection \& Optimization:} Interaction data $\mathcal{D}_1^t$ (without feedback) and $\mathcal{D}_2^t$ (with feedback) are collected to iteratively optimize the model. (c) \textbf{Subsequent Interaction \& Improvement:} When encountering a query involving the similar preference in subsequent sessions $t+k$, the updated assistant $\pi_{\theta_{t+k}}$ successfully incorporates learned preferences to provide a personalized response.}
    \label{fig:schema}
\end{figure*}
\section{Setup}\label{sec:setup}
To effectively evaluate and optimize the personalization capabilities of the assistant model, we define a real-world-motivated interaction scenario for continual personalization. As shown in Figure \ref{fig:schema}, a personalized assistant $\pi_{\theta}$ interacts with a group of $N$ users $U=\{u_1, u_2, \dots, u_N\}$, each with distinct preferences. Each user engages in a total of $T$ dialogue sessions with the assistant, where each session consists of a single-turn interaction. Appendix~\ref{app:multi_turn_extension} details how the same framework can be extended to multi-turn sessions with session-level feedback. Specifically, for the $t$-th session, we denote the current model parameters as $\theta_t$, and user $u_i$ poses a query $q_i^t$ to request task completion, where the task associated with the query belongs to a specific domain in the domain set $\mathcal{X}$. The assistant then generates a response $a_i^t = \pi_{\theta_t}(q_i^t)$.

A key aspect of this scenario is the modeling of latent user preferences and sparse user feedback. Each user $u_i$ is associated with a latent preference profile $\eta_i$, hidden from the assistant, which can be expressed as a concise description of user preferences across domains. When the response $a_i^t$ does not fully align with $\eta_i$, the user experiences a reduced level of satisfaction. To reflect the sparsity of real-world feedback, we introduce a feedback probability $p$: after each interaction, users provide explicit feedback only with probability $p$. When provided, feedback consists of textual feedback $f_i^t$, which provides a qualitative evaluation of the response with respect to the user's preferences, including both positive aspects and deficiencies, and a numerical score $s_i^t$ indicating the overall degree of satisfaction.
The goal of this scenario is to support continuous personalized optimization. At early stages, the assistant lacks knowledge of user preferences and produces weakly personalized responses. By collecting interaction tuples $(q_i^t, a_i^t)$ along with sparse feedback $(f_i^t, s_i^t)$ when available, we apply the optimization procedure described in Section~\ref{sec:methodology} to update the assistant from $\pi_{\theta_t}$ to $\pi_{\theta_{t+1}}$. Over time, the assistant is expected to internalize user preferences and generalize them across related tasks.

\section{Methodology}\label{sec:methodology}
In this section, we elaborate on the optimization phase within the \emph{``Interact-Collect-Optimize''} loop, detailing how the interaction data collected in each session is leveraged to continuously optimize the assistant's personalization performance. Formally, we associate each user $u_i$ with a learnable personalized embedding $\mathbf{v}_i \in \mathbb{R}^{l \times d}$, where $l$ corresponds to the number of virtual tokens allocated to capture user-specific preferences and $d$ denotes the dimensionality of the model's token embeddings.
These embeddings constitute a dynamically maintained collection of user-specific parameters iteratively updated through the \emph{``Interact-Collect-Optimize''} loop, where the set at the beginning of the $t$-th session is denoted as $\mathcal{V}_t = \{\mathbf{v}_1^t, \mathbf{v}_2^t, \dots, \mathbf{v}_N^t\}$, with the initial set $\mathcal{V}_1$ randomly initialized prior to the first session.
Specifically, for session $t$, given the current model parameters $\theta_t$ and the personalized embeddings $\mathcal{V}_t$, the assistant interacts with users and generates responses. We denote $\Omega_1^t$ as users without feedback and $\Omega_2^t$ as users with feedback, such that $|\Omega_1^t| + |\Omega_2^t| = N$, yielding datasets $\mathcal{D}_1^t = \{(q_i^t, a_i^t)\}_{i \in \Omega_1^t}$ and $\mathcal{D}_2^t = \{(q_i^t, a_i^t, f_i^t, s_i^t)\}_{i \in \Omega_2^t}$. Subsequently, we leverage this collected data to update both the model parameters and the personalized embeddings by simultaneously optimizing three distinct objectives within a unified optimization step.

\subsection{Supervised Fine-Tuning for Preference Capture}
To effectively capture the personalized preferences of user $u_i$ while saving context space, we utilize the corresponding personalized embedding $\mathbf{v}_i^t$ to capture the user's implicit preferences $\eta_i$. When valid feedback is available (i.e., for dataset $\mathcal{D}_2^t$), we treat the textual feedback $f_i^t$ as a glimpse into the user's implicit preferences $\eta_i$, manifested as the user's evaluation of the model's response $a_i^t$ to the query $q_i^t$.
Formally, given the query $q_i^t$ and the response $a_i^t$, we prepend the personalized embedding $\mathbf{v}_i^t$ to the sequence of text embeddings and train the personalized embedding to maximize the likelihood of generating the textual feedback $f_i^t$. The goal of this supervised fine-tuning objective is to minimize the following loss:
\begin{equation}\label{eq:sft_loss}
\begin{aligned}
\mathcal{L}_{\text{SFT}}
= - \mathbb{E}_{i \in \Omega_2^t}\Bigg[
&\sum_{k=1}^{|f_i^t|} \log \pi_{\theta_t}\big(
f_{i, k}^t \mid \mathbf{v}_i^t, q_i^t, \big.\\
&\big. a_i^t, f_{i, <k}^t
\big) \Bigg],
\end{aligned}
\end{equation}
where $f_{i, k}^t$ denotes the $k$-th token of the textual feedback $f_i^t$. In this objective, we exclusively calculate gradients for the personalized embedding $\mathbf{v}_i^t$, while treating the model parameters $\theta_t$ as frozen. This forces the preference information contained in the textual feedback to flow into the personalized embedding.

\subsection{Reinforcement Learning for Self-Evaluation Calibration}
The primary obstacle in optimizing personalization is the sparsity of user feedback in real-world scenarios. When explicit feedback is absent, the lack of supervisory signals prevents optimizing the model's personalized responses, resulting in underutilization of valuable interaction data. To address this, we train the assistant model to act as a self-evaluator. Specifically, we perform self-evaluation on the model's response $a_i^t$ given $q_i^t$, with the personalized embedding $\mathbf{v}_i^t$ prepended to the text embeddings to guide this process. The model first generates the self-reflection, speculating on the user's textual feedback, and subsequently predicts the numerical feedback score $\hat{s}_i^t$. To facilitate learning, we introduce a teacher-forcing strategy: with a probability $q$, we include the textual feedback $f_i^t$ in the input context. This exposure allows the model to adjust its evaluation logic to align with the user's actual standards.
We design a composite reward function $R_{\text{eval}}(\cdot)$ consisting of a sparse binary match reward $r_{i,\text{bin}}^t$ and a normalized dense distance-based reward $r_{i,\text{dist}}^t$:
\begin{equation}\label{eq:r_eval}
    R_{\text{eval}}(\hat{s}_i^t, s_i^t) = r_{i,\text{bin}}^t + r_{i,\text{dist}}^t,
\end{equation}
where the two components are calculated as:
\begin{equation}
\begin{aligned}
    r_{i,\text{bin}}^t = 
    \begin{cases} 
    1 ,& \text{if } \hat{s}_i^t = s_i^t \\ 
    0, & \text{otherwise} 
    \end{cases} \\
    r_{i,\text{dist}}^t = \frac{C_{\max} - |\hat{s}_i^t - s_i^t|}{C_{\max}}.
\end{aligned}
\end{equation}
Here, $C_{\max}$ denotes the maximum possible score value (i.e., the upper bound of the score range). We incorporate $r_{i,\text{dist}}^t$ alongside the binary reward because the binary signal alone is too sparse, especially early in training when predictions are inaccurate, providing little learning signal for effective policy optimization. The distance-based term, normalized by $C_{\max}$ to align its scale with $r_{i,\text{bin}}^t$, supplies a dense reward signal that guides the predicted score toward the true score, while $r_{i,\text{bin}}^t$ encourages precise alignment once predictions are sufficiently close. The goal of this reinforcement learning objective is to maximize the expected reward defined as:
\begin{equation}\label{eq:j_eval}
    \mathcal{J}_{\text{eval}} = \mathbb{E}_{i \in {\Omega}_2^t} \left[ R_{\text{eval}}(\hat{s}_i^t, s_i^t) - \beta \text{KL}(\pi_{\theta_t} || \pi_{\text{ref}}) \right],
\end{equation}
where $\pi_{\text{ref}}$ is the original policy without any optimization (i.e., $\pi_{\theta_1}$). In this objective, we compute gradients for both the personalized embeddings and the assistant model: backpropagation refines $\mathbf{v}_i^t$ to better capture user preferences, while simultaneously calibrating the model to score its own responses according to user standards. As a result, for users whose prior feedback has helped calibrate self-evaluation, the model can generate proxy rewards via self-evaluation when explicit feedback is unavailable in subsequent interactions, mitigating feedback sparsity, maximizing the utilization of all interaction data, and improving learning efficiency.
\subsection{Reinforcement Learning for Personalized Response Optimization}
This objective focuses on optimizing the generation of response $a_i^t$ given the query $q_i^t$ under the guidance of personalized embeddings, addressing sparse feedback by leveraging the self-evaluation capabilities established in the preceding \emph{``Interact-Collect-Optimize''} loop. To prevent the degradation of the model's general capabilities, we define a composite reward composed of two distinct factors: \emph{completeness} and \emph{personalization}.
First, to assess whether the assistant's response fulfills the task requirements specified by the user query, we employ an extra evaluator (implemented via Qwen-Flash~\citet{alibaba2026qwenflash}). This evaluator provides a binary completeness reward $r_{i,\text{comp}}^t \in \{0, 1\}$, where $1$ indicates that the response is sufficiently comprehensive to be considered complete and $0$ indicates missing key points. This evaluation focuses strictly on task completion and is independent of user preference alignment.
Second, for the personalization component, we utilize the hybrid signal derived from real or proxy feedback. We denote this personalization score as $r_{i,\text{pers}}^t$:
\begin{equation}
    r_{i,\text{pers}}^t = 
    \begin{cases} 
        \hat{s}_i^t,  & \text{if } i \in \Omega_1^t \quad (\text{Proxy Feedback})\\
        s_i^t, & \text{if } i \in \Omega_2^t \quad (\text{Real Feedback}) 
    \end{cases}.
\end{equation}
The final reward $r_i^t$ is the product: $r_i^t = r_{i,\text{comp}}^t \cdot r_{i,\text{pers}}^t$, ensuring the model prioritizes completeness as a prerequisite for personalization, as responses that fail the task receive no policy reward even if they match user preferences, reducing the incentive to trade off task completion for personalized style.
We then define the goal of this reinforcement learning objective as maximizing the following expected reward:
\begin{equation}\label{eq:j_policy}
    \mathcal{J}_{\text{policy}} = \mathbb{E}_{i \in {\Omega}_1^t \cup {\Omega}_2^t} \left[ r_i^t - \beta \text{KL}(\pi_{\theta_t} || \pi_{\text{ref}}) \right],
\end{equation}
where $\pi_{\text{ref}}$ is the original policy without any optimization (i.e., $\pi_{\theta_1}$). In this objective, we calculate gradients exclusively for the model parameters $\theta_t$, treating the personalized embedding $\mathbf{v}_i^t$ as frozen. This encourages the model to utilize the preference representations captured in personalized embedding to generate personalized responses.

\subsection{Synergistic Updating}
Having established the three optimization objectives above, to prevent off-policy issues caused by sequential optimization of these objectives, we aggregate the gradients from all three objectives to update the model and personalized embedding parameters simultaneously. Specifically, the gradients for Preference Capture (SFT) are derived via backpropagation from the loss in Eq.~\ref{eq:sft_loss}. Meanwhile, the gradients for Self-Evaluation Calibration (RL) and Personalized Response Optimization (RL) are derived by maximizing the objectives shown in Eq.~\ref{eq:j_eval} and Eq.~\ref{eq:j_policy}, respectively, via the Proximal Policy Optimization (PPO) algorithm~\citet{schulman2017proximal} (detailed in Appendix~\ref{app:ppo}). Finally, we aggregate the gradients of these three objectives to perform the parameter updates. This ensures that both Self-Evaluation Calibration and Personalized Response Optimization are optimized using data sampled from the current policy (on-policy), thereby stabilizing the learning process. For clarity, the overall optimization process is summarized in Algorithm~\ref{alg:optimization}, with a schematic illustration of the personalized embedding integration and objective computation provided in Appendix~\ref{app:pvec}.

\section{Experiments and Results}
\label{sec:experiments}
\subsection{Experimental Setup}
Considering user privacy protection and the difficulty of collecting real-world interaction data, we employ the PersonaLens~\citet{zhao-etal-2025-personalens} dataset to instantiate the personalized interaction scenario defined in Section~\ref{sec:setup}. PersonaLens is a public personalization benchmark covering $20$ domains (i.e., $|\mathcal{X}| = 20$), where each domain contains $4$ to $5$ tasks centered on the user's preferences within that domain, and all preference types can be fully exposed through a Minimal Task Set (MTS) of no more than $3$ tasks. PersonaLens also pre-defines a set of interested domains per user (ranging from $9$ to $20$), reflecting that users are not necessarily interested in all domains. To balance domain and user diversity, we select $256$ users, each interested in exactly $17$ domains, which together cover all $20$ domains.

We further standardize task configurations by fixing the number of tasks per domain to $4$, randomly discarding one non-MTS task in $5$-task domains to preserve complete preference exposure. The interaction order is strictly controlled: for each user, the assistant traverses interested domains in alphabetical order, and within each domain, tasks are sampled without replacement with MTS tasks prioritized before the remaining ones. This design guarantees that (1) all preference types within a domain are exposed after the first three traversals, and (2) during the fourth traversal, the remaining task involves only previously observed preferences (illustrated in Appendix~\ref{app:traversal_protocol}). Accordingly, we perform continual optimization during the first three traversals, updating the model after each of the $51$ interactions ($3$ tasks across $17$ domains), and reserve the fourth traversal for evaluation. This chronological split ensures that test queries involve already exposed preferences but unseen tasks, enabling a controlled assessment of continual personalization. Performance is measured by averaging the product of the personalization score and a binary completeness score over all users, domains, and interactions, reported as mean$\pm$std over three runs; Appendix~\ref{app:decomposed_main_results} reports the two component scores separately.

\subsection{Comparison Methods}
To validate the effectiveness of our proposed COPE method, we compare it against several representative baselines. We first include \textbf{Base}, where the assistant directly interacts with the user without additional optimization. For training-free personalization, we compare with Profile-Augmented Prompting (\textbf{PAP}) and Retrieval-Augmented Prompting (\textbf{RAP}), including Sparse Retrieval (\textbf{SR}), Dense Retrieval (\textbf{DR}), and a query-expansion variant of dense retrieval (\textbf{Q-DR}). For training-based personalization, we include Training Profile-Augmented Prompting (\textbf{T-PAP}), which serves to isolate the effect of training from the use of learnable personalized embeddings. In addition, to examine whether COPE is complementary to training-free methods, we report the performance of the COPE-optimized model integrated with SR, denoted as \textbf{COPE+SR}. We also integrate T-PAP with the same SR module (\textbf{T-PAP+SR}) to control for the contribution of retrieval. We evaluate all methods under three feedback probability settings ($p \in \{0.25, 0.50, 0.75\}$). Implementation details of COPE and the comparison methods are provided in Appendix~\ref{app:implementation_details}. The experimental results are presented in Table~\ref{tab:main_results}.

\begin{table}[t]
    \centering
    \caption{Scores (personalization score $\times$ binary completeness score, out of $10$) of models optimized by different methods under varying user feedback probabilities ($\text{mean}\pm\text{std}$). 
    $\dagger$ indicates methods that involve training.}
    \label{tab:main_results}
    \setlength{\tabcolsep}{3pt}
    \begin{tabular}{@{}l c c c@{}}
        \toprule
        \multirow{2}{*}{\textbf{Methods}} & \multicolumn{3}{c}{\textbf{User Feedback Probability}} \\
        \cmidrule(lr){2-4} 
         & $p=0.25$ & $p=0.50$ & $p=0.75$ \\ \hline
        Base & $2.36{\pm}0.00$ & $2.36{\pm}0.00$ & $2.36{\pm}0.00$ \\ \hline
        SR & $3.08{\pm}0.03$ & $3.09{\pm}0.01$ & $3.11{\pm}0.04$ \\ \hline
        DR & $3.07{\pm}0.05$ & $3.15{\pm}0.03$ & $3.25{\pm}0.02$ \\ \hline
        Q-DR & $3.15{\pm}0.07$ & $3.36{\pm}0.06$ & $3.47{\pm}0.06$ \\ \hline
        PAP  & $3.57{\pm}0.03$ & $3.65{\pm}0.02$ & $3.69{\pm}0.01$ \\ \hline
        T-PAP$^\dagger$ & $4.45{\pm}0.04$ & $4.47{\pm}0.01$ & $4.69{\pm}0.02$ \\ \hline
        COPE$^\dagger$  & $4.58{\pm}0.06$ & $4.69{\pm}0.05$ & $4.74{\pm}0.02$ \\ \hline
        T-PAP$^\dagger$+SR & $\underline{4.76{\pm}0.02}$ & $\underline{4.83{\pm}0.04}$ & $\underline{4.98{\pm}0.03}$ \\ \hline
        COPE$^\dagger$+SR & $\textbf{5.08}{\pm}0.03$ & $\textbf{5.23}{\pm}0.05$ & $\textbf{5.36}{\pm}0.02$ \\ 
        \bottomrule
    \end{tabular}
\end{table}

\subsection{Main Results}
As presented in Table \ref{tab:main_results}, performance generally improves from training-free to training-based personalization, while combining training with retrieval yields the strongest results, indicating that both richer inference-time context and training-based adaptation improve personalization.
Among training-free methods, approaches using stronger inference-time signals perform better: DR generally improves over SR, Q-DR further benefits from query expansion, and PAP achieves the best training-free results by summarizing user history into a coherent profile, though at the cost of additional context and latency.
COPE substantially improves over these baselines; for example, at $p=0.75$, it reaches $4.74$ compared with $3.69$ for PAP.
Its advantage over T-PAP, which uses the same reinforcement learning objectives but no learnable personalized embeddings, shows that the gain is not merely due to training but also to the personalized embedding representation.
Finally, COPE+SR achieves the best scores in all settings, peaking at $5.36$ when $p=0.75$, confirming that learned implicit preferences and retrieved explicit histories provide complementary personalization signals. Adding SR improves T-PAP by $0.29$--$0.36$ points, but improves COPE by a larger $0.50$--$0.62$ points. This suggests that T-PAP's textual summary partially overlaps with retrieved histories, whereas COPE's compact latent preferences are more complementary to explicit retrieved evidence; accordingly, COPE+SR consistently outperforms T-PAP+SR by $0.32$--$0.40$ points.

\begin{figure}[ht]
    \centering
    \includegraphics[width=1\linewidth]{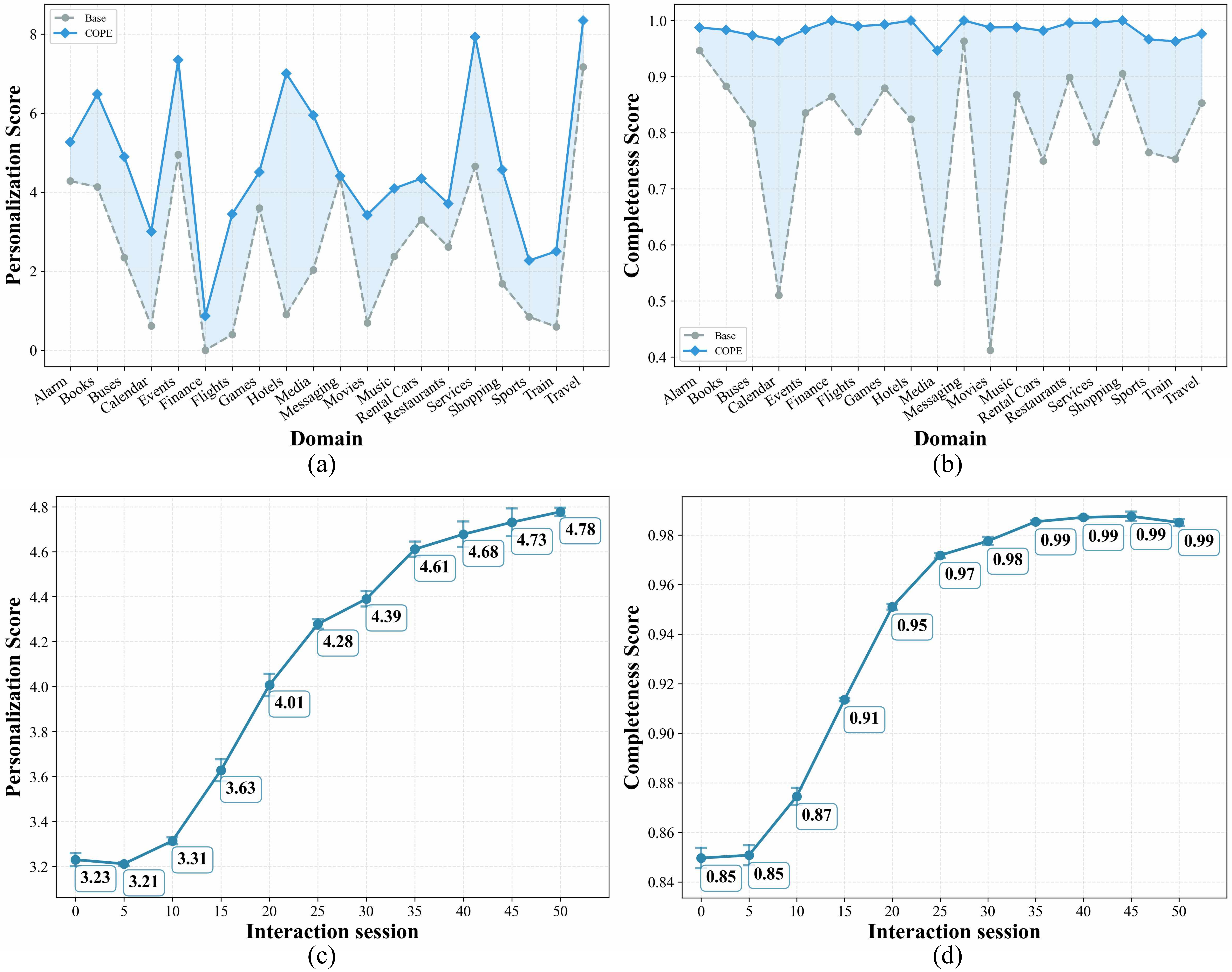}
    \caption{Analysis of COPE performance. (a-b) Comparison of personalization and completeness scores against Base across 20 domains. (c-d) Training dynamics of personalization and completeness scores over interaction sessions.}
    \label{fig:ablations1}
\end{figure}
\begin{figure}[ht]
    \centering
    \includegraphics[width=1\linewidth]{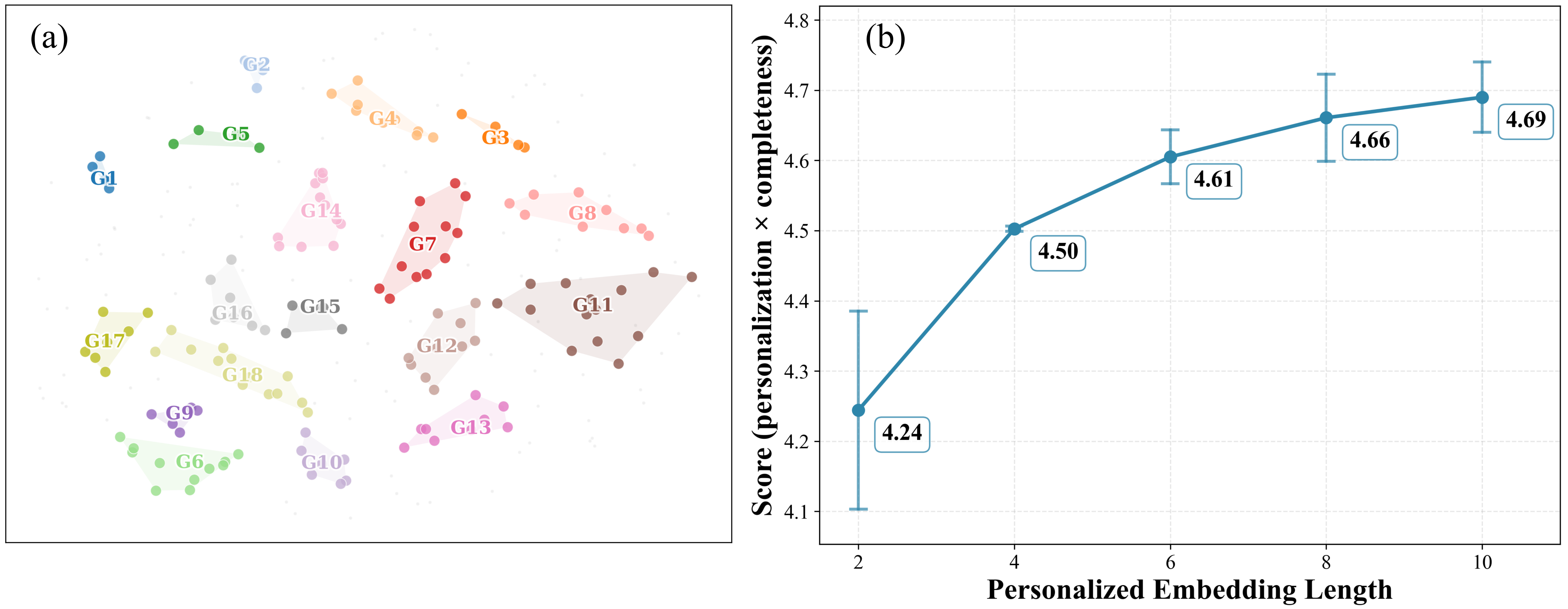}
    \caption{ (a) Visualization of personalized embedding clusters. (b) Impact of personalized embedding length $l$ on COPE.}
    \label{fig:ablations2}
\end{figure}
\subsection{Ablation and Analysis}
\textbf{Performance Dynamics.} Figure~\ref{fig:ablations1}(a-b) compares COPE with Base on personalization and completeness across all $20$ domains, showing consistent gains and completeness scores near 1.0. Across sessions (Figure~\ref{fig:ablations1}(c-d)), completeness rises rapidly within the first $20$ sessions and then stabilizes, whereas personalization improves more gradually before tapering off as additional history provides diminishing marginal information.
To examine whether the self-evaluation signal aligns with user feedback, we evaluate the assistant-estimated and user-provided personalization scores at two granularities. At the population level, the Pearson correlation between user-averaged trajectories reaches $r=0.81$ during training and $r=0.71\pm0.06$ on three held-out runs. At the instance level, a pairwise ranking analysis on the validation tasks (Appendix~\ref{app:instance_level_self_eval}) shows that the proxy ranks consistently with the user on $72.63\%$ of pairs where both produce a directional judgment. These results indicate that the self-evaluation signal tracks user feedback at both the population and individual interaction level. Appendix~\ref{app:self_eval_mechanism} provides mechanistic intuition and qualitative examples.

\textbf{Embedding Visualization.} To interpret the learned embeddings, we cluster $\mathcal{V}_{50}$ using UMAP \citet{mcinnes2018umap} and HDBSCAN \citet{campello2013density} (Figure \ref{fig:ablations2}(a)). Two adjacent clusters, $\text{G}1$ and $\text{G}2$, illustrate how spatial relation aligns with preference patterns: both contain high-income users who prefer business or first-class travel, yet they separate by finer sub-preferences, with $\text{G}1$ favoring modern aesthetics (e.g., electronic music, competitive gaming) and $\text{G}2$ leaning toward traditional tastes (e.g., classical music, chess). This example suggests that the embeddings organize users according to preference patterns beyond coarse demographics.

\begin{table}[t]
    \centering
    \caption{Ablation of optimization objectives in COPE ($p=0.5$).}
    \label{tab:objective_ablation}
    \begin{tabular}{c c c c}
        \toprule
        $\mathcal{L}_{\text{SFT}}$ & $\mathcal{J}_{\text{eval}}$ & $\mathcal{J}_{\text{policy}}$ & Scores \\
        \midrule
        \ding{55} & \ding{55} & \ding{51} & $3.36\pm0.06$\\ \hline
        \ding{55} & \ding{51} & \ding{51} & $4.57\pm0.04$\\ \hline
        \ding{51} & \ding{51} & \ding{51} & $\textbf{4.69}\pm0.05$\\

        \bottomrule
    \end{tabular}
\end{table}
\begin{table}[t]
    \centering
    \caption{Ablation of trainable parameter configurations for each objective in COPE ($p=0.5$).}
    \label{tab:parameters_ablation}
    \begin{tabular}{c c c c}
        \toprule
        $\mathcal{L}_{\text{SFT}}$ & $\mathcal{J}_{\text{eval}}$ & $\mathcal{J}_{\text{policy}}$ & Scores \\
        \midrule
        $\mathbf{v}, \theta$ & $\mathbf{v}, \theta$ & $\theta$ & $4.18\pm0.07$\\ \hline
        $\mathbf{v}$ & $\mathbf{v}, \theta$ & $\mathbf{v}, \theta$ & $4.45\pm0.08$\\ \hline
        $\mathbf{v}$ & $\theta$ & $\theta$ & $4.49\pm0.07$\\ \hline
        $\mathbf{v}$ & $\mathbf{v}$ & $\theta$ & $4.53\pm0.05$\\ \hline
        $\mathbf{v}$ & $\mathbf{v}, \theta$ & $\theta$ & $\textbf{4.69}\pm0.05$\\

        \bottomrule
    \end{tabular}
\end{table}

\textbf{Embedding Length.} The capacity of personalized embeddings is governed by their length $l$. Figure \ref{fig:ablations2}(b) illustrates the performance of COPE ($p=0.5$) across different $l$ values. While performance improves with $l$, it exhibits diminishing marginal returns beyond $l=4$. Balancing performance maximization with parameter efficiency, we select $l=10$ as the final configuration. Under this setting, the user-specific parameters introduce an overhead of only approximately $1.2\times10^{-5}$ of the total LLM (Qwen3-1.7B~\citet{yang2025qwen3}) parameters $\theta$ per user, ensuring the method remains highly efficient for large-scale deployment.

\textbf{Objective Ablation.} We analyze the contribution of each optimization objective in Table \ref{tab:objective_ablation}. Removing either $\mathcal{L}_{\text{SFT}}$ or $\mathcal{J}_{\text{eval}}$ degrades performance, with the omission of $\mathcal{J}_{\text{eval}}$ having the most pronounced impact. Without $\mathcal{L}_{\text{SFT}}$ and $\mathcal{J}_{\text{eval}}$, the model fails to explicitly leverage textual and numerical feedback, respectively. Consequently, the personalized embeddings fail to effectively capture user preferences, and the model is constrained to learning exclusively from the subset of data containing explicit feedback ($\mathcal{D}_2$), thereby substantially curtailing learning efficiency.

\textbf{Trainable parameter configuration Ablation.} Table \ref{tab:parameters_ablation} explores the impact of updating different parameter sets (personalized embeddings $\mathbf{v}$ and LLM weights $\theta$) for each objective. For $\mathcal{L}_{\text{SFT}}$, additionally updating $\theta$ degrades performance ($4.18$), since this objective focuses on injecting textual preference knowledge into $\mathbf{v}$, and updating $\theta$ likely disrupts the LLM's general capabilities. For $\mathcal{J}_{\text{eval}}$, jointly updating $\mathbf{v}$ and $\theta$ is optimal, as this objective must simultaneously compress preferences into $\mathbf{v}$ and calibrate the scoring capability within $\theta$, and updating only one leads to suboptimal alignment. For $\mathcal{J}_{\text{policy}}$, training only $\theta$ is most effective, since the model here utilizes the captured personalized information to generate responses; forcing updates on $\mathbf{v}$ causes overfitting as no new preference information is introduced, leading to a performance drop ($4.45$).

\section{Discussion}
\textbf{General Capability Retention.} Since continual personalization updates the shared model parameters, a natural concern is whether it degrades the model's general capabilities. To examine this issue, we evaluate the model after the full optimization phase, i.e., after updating the model across the $51$ interactions from the first three traversals ($3$ tasks across $17$ domains), and compare it with the original Qwen3-1.7B~\citet{yang2025qwen3}. As shown in Appendix~\ref{app:general_capability}, the optimized model maintains general capabilities after continual personalization. These results indicate that COPE improves personalization without sacrificing the model's general capability, suggesting that the proposed continual optimization does not induce catastrophic forgetting over the evaluated training horizon.

\begin{figure}[!t]
    \centering
    \begin{subfigure}{0.95\linewidth}
        \centering
        \includegraphics[width=\linewidth]{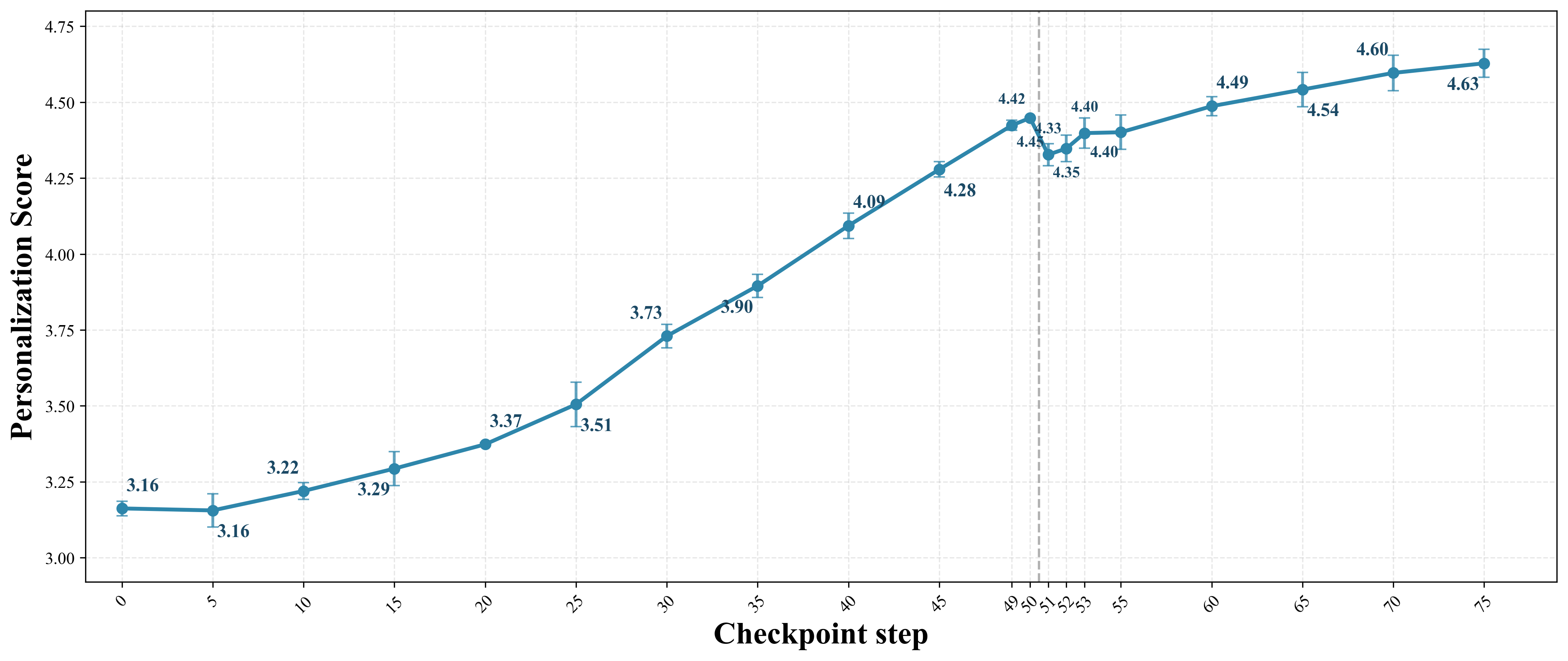}
        \caption{Personalization score.}
        \label{fig:user_shift_personalization}
    \end{subfigure}
    \vspace{0.4em}
    \begin{subfigure}{0.95\linewidth}
        \centering
        \includegraphics[width=\linewidth]{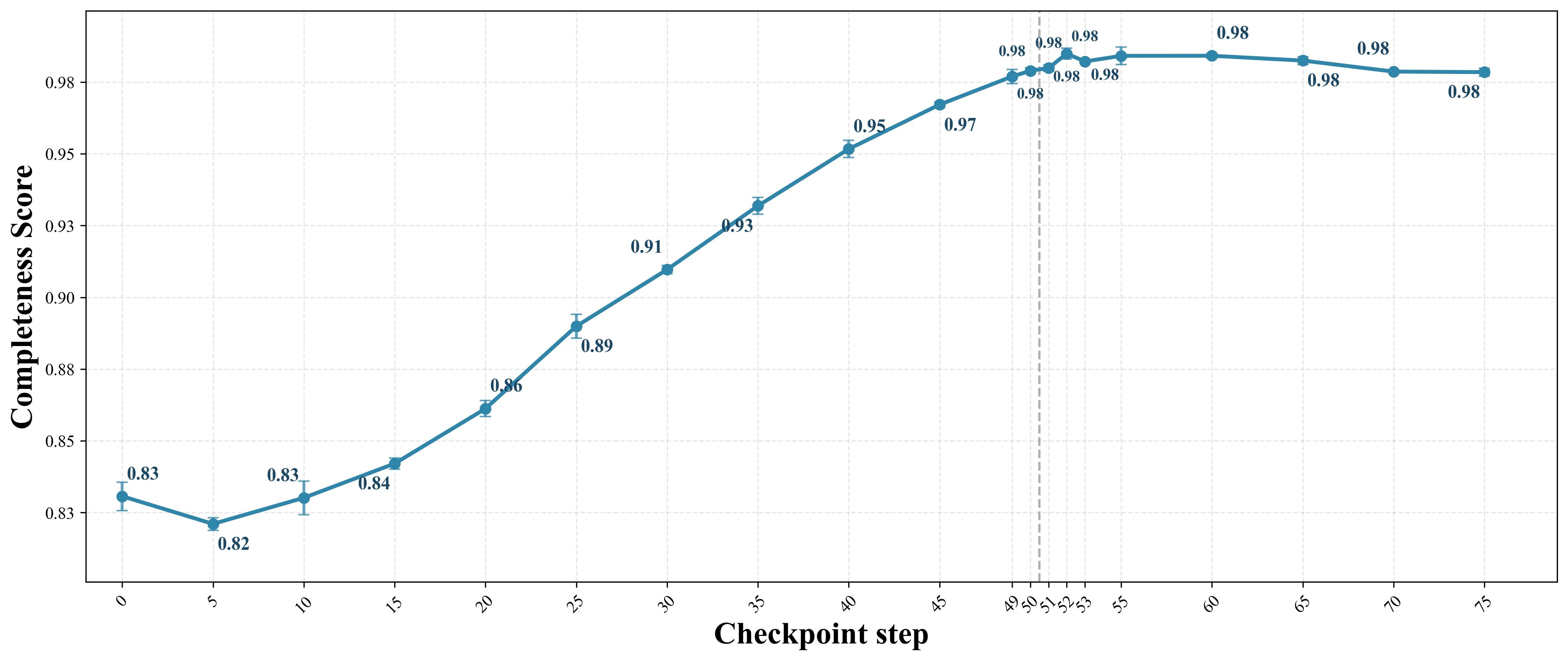}
        \caption{Completeness score.}
        \label{fig:user_shift_completeness}
    \end{subfigure}
    \caption{Within-domain preference shift evaluation. The vertical dashed line marks the simulated preference shift. Personalized embeddings are carried over rather than re-initialized after the shift.}
    \label{fig:user_preference_shift}
\end{figure}

\begin{figure}[!t]
    \centering
    \begin{subfigure}{0.95\linewidth}
        \centering
        \includegraphics[width=\linewidth]{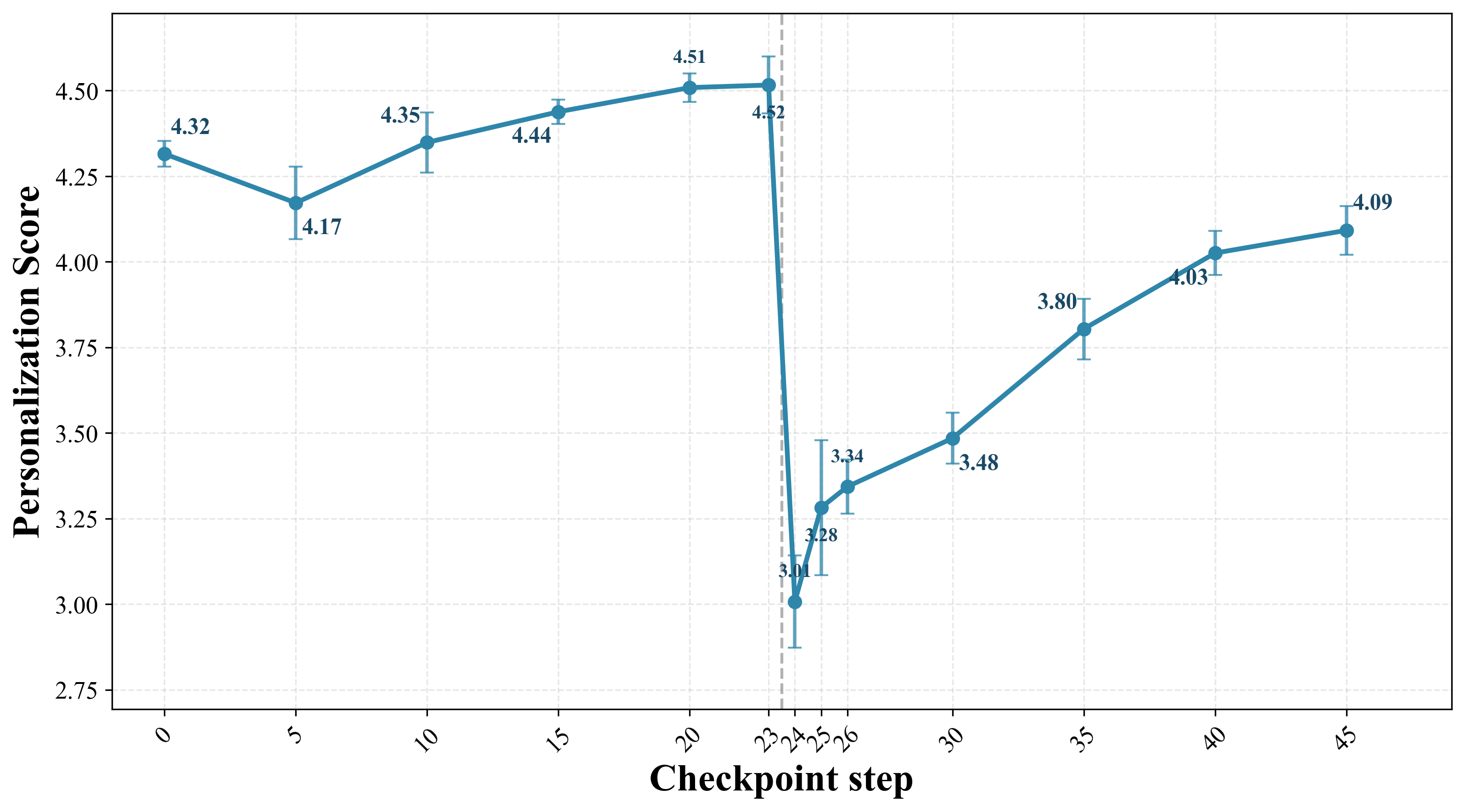}
        \caption{Personalization score.}
        \label{fig:domain_shift_personalization}
    \end{subfigure}
    \vspace{0.4em}
    \begin{subfigure}{0.95\linewidth}
        \centering
        \includegraphics[width=\linewidth]{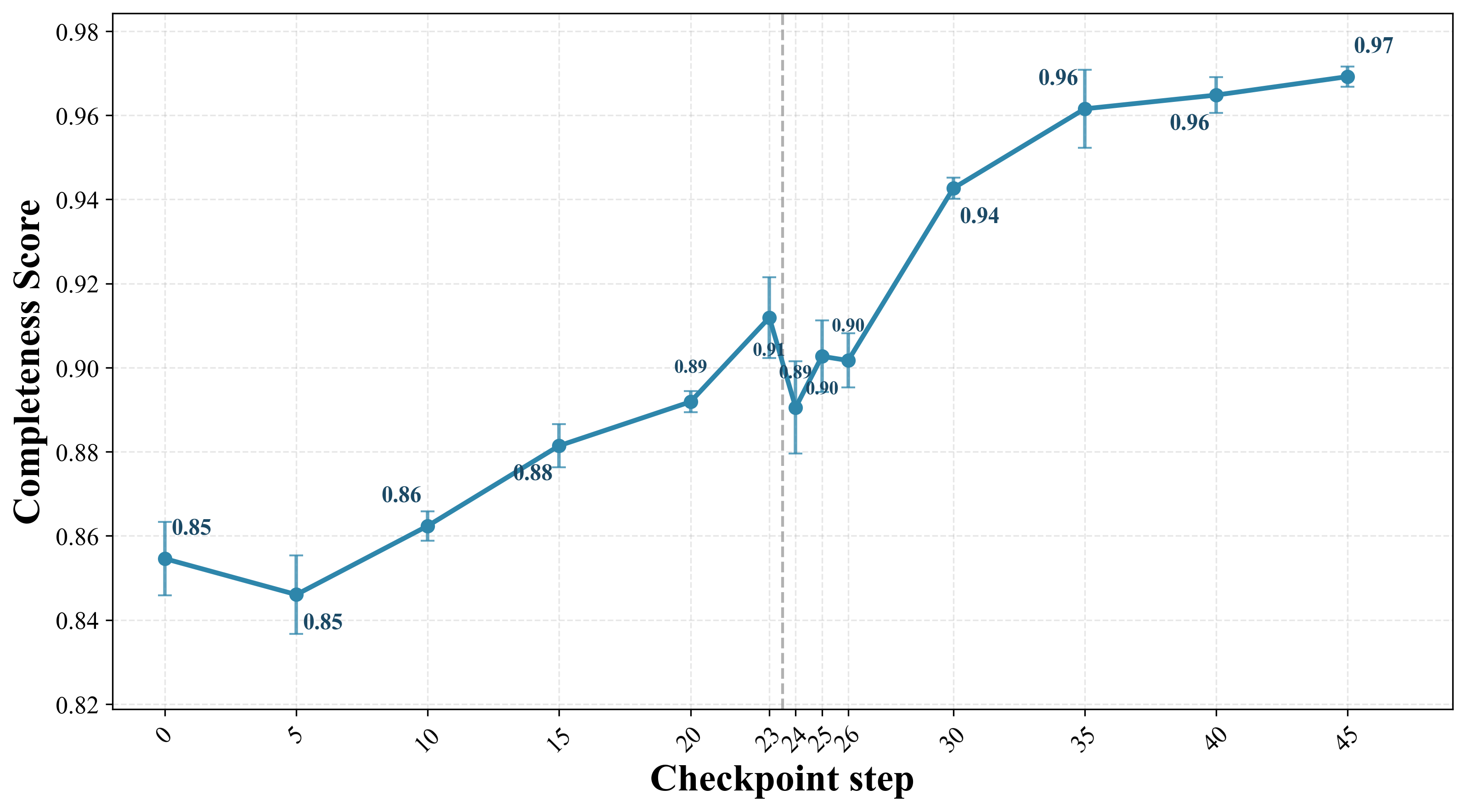}
        \caption{Completeness score.}
        \label{fig:domain_shift_completeness}
    \end{subfigure}
    \caption{New-domain preference shift evaluation. The vertical dashed line marks the simulated preference shift. Personalized embeddings are carried over rather than re-initialized after the shift.}
    \label{fig:domain_preference_shift}
\end{figure}

\textbf{Adaptation to Preference Shifts.} In realistic deployments, user preferences may evolve rather than remain fixed. To examine whether COPE can adapt under such distribution shifts, we conduct two simulated preference-shift experiments in which the personalized embeddings are carried over across the shift rather than re-initialized. The first simulates shifted preferences within previously seen domains, while the second imposes a more stringent shift to non-overlapping new domains. As shown in Figures~\ref{fig:user_preference_shift} and~\ref{fig:domain_preference_shift}, COPE maintains or quickly recovers both personalization and completeness under the within-domain preference shift, and it also rapidly adapts under the more challenging new-domain preference shift despite an immediate post-shift drop. Experimental details are provided in Appendix~\ref{app:preference_shift}.

\textbf{Scaling to Larger Models.} To examine whether COPE scales beyond the Qwen3-1.7B backbone used in our main experiments, we conduct additional experiments with Qwen3-4B, Qwen3-8B, and Llama-3.1-8B-Instruct~\citet{grattafiori2024llama} under $p=0.5$. Table~\ref{tab:model_scaling} retains the original Qwen3-1.7B result for reference. COPE consistently improves over Base across all backbones, while its score within the Qwen3 family increases monotonically from $4.69$ at 1.7B to $5.49$ at 8B, demonstrating improved performance with greater backbone capacity. The improvement on Llama-3.1-8B-Instruct shows that COPE remains effective with a non-Qwen assistant, thereby alleviating concerns about same-family coupling between the assistant and the Qwen-Flash simulator and completeness evaluator.

\begin{table}[t]
    \centering
    \caption{Model-scaling results with $p=0.5$. Base denotes the corresponding unoptimized model, and COPE denotes the model after continual personalization training.}
    \label{tab:model_scaling}
    {
    \small
    \setlength{\tabcolsep}{3pt}
    \begin{tabular}{lcc}
        \toprule
        Backbone & Base & COPE \\
        \midrule
        Qwen3-1.7B & $2.36\pm0.00$ & $4.69\pm0.05$ \\
        Qwen3-4B & $3.17\pm0.00$ & $5.26\pm0.04$ \\
        Qwen3-8B & $3.60\pm0.01$ & $5.49\pm0.03$ \\
        Llama-3.1-8B-Instruct & $3.20\pm0.04$ & $5.16\pm0.06$ \\
        \bottomrule
    \end{tabular}}
\end{table}

\textbf{Robustness to Evaluator Choice.} To examine whether COPE's gains depend on the Qwen-family pipeline shared by the user simulator and the rubric-based scoring across training and evaluation, we conduct two complementary checks. First, swapping the user simulator with DeepSeek-V3~\citet{liu2024deepseek} under $p=0.5$ and $q=0.5$ yields lower absolute scores but preserves the relative ordering, with COPE still achieving the best performance (Appendix~\ref{app:user_model_robustness}). Second, we conduct a blinded pairwise comparison between COPE and T-PAP on $204$ aligned validation response pairs using a human annotator; the annotator does not receive the simulator's rubrics or numerical scores. Results show that the human annotator consistently prefers COPE over T-PAP (Figure~\ref{fig:third_party_judge}; Appendix~\ref{app:third_party_judge}). Both checks indicate that COPE remains effective under different user-model judging styles and that the observed gains are not driven by closed-loop bias.

\section{Conclusion}
In this work, we proposed COPE, a unified framework that enables continual personalization optimization with learnable embeddings and self-evaluation. Experiments demonstrate that COPE improves personalization under sparse feedback, remains complementary to retrieval-based prompting, and learns personalized embeddings that capture meaningful preference patterns. Additional analyses show that COPE's self-evaluation aligns with user feedback, the optimized model largely preserves general capabilities and recovers from preference shifts within seen domains and across new domains, and the gains remain robust under alternative evaluators, suggesting that COPE is a practical step toward continually adaptive personalized assistants. Thanks to the complementarity between COPE and retrieval-augmented prompting, a deployment with millions of users can update one user subset at a time while supporting the remaining users through retrieval over their interaction histories, without sacrificing personalization support for users awaiting parametric updates. Future work will aim to extend the framework to more complex multi-domain multi-turn tasks and privacy-preserving evaluations in real-user scenarios.

\section*{Limitations}
Due to user-privacy considerations and the limited availability of longitudinal real user--assistant interaction data, our current evaluation uses benchmark-derived simulated users from PersonaLens. This design enables controlled and reproducible evaluation under sparse feedback without collecting sensitive real-user interaction traces. Future work can complement this controlled evidence with privacy-preserving real-user evaluation and additional datasets. Our chronological protocol already spans $51$ interactions per user during optimization; evaluating COPE over still longer horizons would provide a valuable further stress test. The current evaluation is also limited to single-turn sessions. Appendix~\ref{app:multi_turn_extension} provides a concrete extension of COPE to multi-turn sessions with session-level feedback, while empirical validation on privacy-preserving multi-turn data remains future work.

\section*{Acknowledgments}
This work was supported by Qwen Business Unit through Alibaba Research Intern Program.

\bibliography{references}

\clearpage
\appendix
\section{Related Work and Positioning}
\label{app:related_work}
Recent work has explored user embeddings as a compact interface between user histories and LLMs. User-LLM~\citep{ning2025user} generates user embeddings from interaction histories and integrates them with LLMs through cross-attention, improving efficiency over text-only user-context prompts. Embedding-to-Prefix~\citep{huber2025embedding} maps dynamically updated user embeddings into a soft prefix for a frozen LLM, providing a lightweight mechanism for continual personalization without updating the LLM backbone. These methods show that compact user representations can reduce context overhead and provide useful personalization signals. COPE shares this motivation but differs in how the user representation is acquired and used: rather than relying on precomputed user encoders or frozen-backbone prefix injection, COPE learns per-user embeddings from sparse textual and numerical feedback inside the interaction loop and uses them together with policy optimization.

Other recent methods study continual or feedback-based personalization. SPRInG~\citep{kim2026spring} combines selective parametric adaptation with retrieval-interpolated generation to handle preference drift and mitigate forgetting in continual personalization. C-BPO~\citep{ma2026personalizing} studies binary-signal preference optimization by treating target-user histories as positive signals and other users' histories as implicit negatives under a preference-corrected objective. These directions are complementary to COPE: they focus on drift-aware semi-parametric adaptation or binary-signal preference optimization, while COPE focuses on sparse-feedback user--assistant interactions where missing feedback is addressed through a self-evaluation-based proxy reward and a unified update step for preference capture, self-evaluation calibration, and personalized response optimization.

\section{Preliminaries on Proximal Policy Optimization (PPO)}\label{app:ppo}
In this work, we employ Proximal Policy Optimization (PPO)~\citep{schulman2017proximal}, a widely adopted policy gradient method, to optimize both the Self-Evaluation Calibration and Personalized Response Optimization objectives. PPO operates within the \textbf{Actor-Critic} framework, consisting of a policy network (Actor) $\pi_{\theta}$ that generates actions (responses) and a value network (Critic) $V_{\phi}$ that estimates the expected return of a given state (query and context).

\paragraph{Importance Sampling.}
PPO is an on-policy algorithm that alternates between data collection and policy optimization. In each iteration, a batch of trajectories is first collected using the current policy, referred to as the old policy $\pi_{\theta_{\text{old}}}$. To enable multiple epochs of optimization over the same batch while accounting for the discrepancy between the updated policy and the data-collecting policy, PPO employs importance sampling. We define the probability ratio (importance weight) as:
\begin{equation}
    r_t(\theta) = \frac{\pi_{\theta}(a_t|s_t)}{\pi_{\theta_{\text{old}}}(a_t|s_t)}.
\end{equation}

\paragraph{Generalized Advantage Estimation (GAE).}
To reduce the variance of policy gradient estimates, PPO uses the Critic $V_{\phi}$ to estimate the advantage using Generalized Advantage Estimation (GAE)~\citep{schulman2015high}. The advantage $\hat{A}_t$ measures how much better an action is compared to the average expectation. It is calculated as:
\begin{equation}
    \begin{aligned}
    \hat{A}_t
    &= \sum_{k=0}^{\infty} (\gamma \lambda)^k \delta_{t+k},\\
    \text{where } \delta_t
    &= r_t + \gamma V_{\phi}(s_{t+1}) - V_{\phi}(s_t).
    \end{aligned}
\end{equation}
Here, $r_t$ denotes the scalar reward at timestep $t$, $\gamma$ is the discount factor, $\lambda$ is the GAE smoothing parameter, and $\delta_t$ represents the temporal difference (TD) error.

\paragraph{Value Network Training.}
In our implementation, we maintain two separate critics: $V_{\phi_{\mathrm{eval}}}$ for Self-Evaluation Calibration and $V_{\phi_{\mathrm{resp}}}$ for Personalized Response Optimization. They are initialized from the same assistant model but trained independently using the returns of their corresponding PPO objective. For $o\in\{\mathrm{eval},\mathrm{resp}\}$, the critic minimizes the value loss
\begin{equation}\label{eq:critic_loss}
\mathcal L_V^o(\phi_o)
=\frac{1}{2}\mathbb E_t
\left[(V_{\phi_o}(s_t)-\hat R_t^o)^2\right],
\end{equation}
where $\hat R_t^o$ is the return estimated by GAE; in our terminal-reward setting with $\gamma=\lambda=1$, it equals the corresponding final reward, i.e., $R_{\mathrm{eval}}(\hat{s}_i^t,s_i^t)$ for Self-Evaluation Calibration and $r_i^t$ for Personalized Response Optimization. Critic gradients are optimized separately and are not included in the joint update of the assistant model and personalized embeddings.

\paragraph{Optimization Objective.}
The core of PPO is to maximize a surrogate objective while constraining the policy update to avoid excessive deviation from the behavior policy. In addition to the clipping mechanism, and following standard practice in RLHF-style optimization, we further incorporate a KL divergence penalty with respect to a fixed reference policy $\pi_{\text{ref}}$ to prevent the learned policy from drifting too far from the pretrained model. The overall objective is defined as:
\begin{equation} \label{eq:ppo_objective}
    \begin{aligned}
    \mathcal{J}_{\text{PPO}}(\theta)
    &= \mathbb{E}_t \Big[
    \mathcal{J}^{\text{CLIP}}_t(\theta)\\
    &\quad {}- \beta \, \mathrm{KL}\big(
    \pi_{\theta}(\cdot|s_t)
    \,\|\, \pi_{\text{ref}}(\cdot|s_t)
    \big)
    \Big].
    \end{aligned}
\end{equation}
Specifically, the clipped surrogate objective is given by:
\begin{equation}
    \begin{aligned}
    \mathcal{J}^{\text{CLIP}}_t(\theta)
    &= \min\!\Big(
    r_t(\theta)\hat{A}_t,\\
    &\quad
    \mathrm{clip}\!\left(
    r_t(\theta), 1-\epsilon, 1+\epsilon
    \right)\hat{A}_t
    \Big),
    \end{aligned}
\end{equation}
where $\epsilon$ is a hyperparameter (e.g., $0.2$) controlling the trust region, and $\beta$ determines the strength of the KL regularization.

\section{Optimization Algorithm}
\label{app:optimization_algorithm}
Algorithm~\ref{alg:optimization} summarizes the complete ``Interact-Collect-Optimize'' loop of COPE, including sparse feedback collection, the computation of the three optimization objectives, and the joint update of model parameters and personalized embeddings. The two PPO objectives use separate critics initialized from the assistant model. Each critic is trained on the returns of its corresponding trajectories and provides the advantages for that objective's policy update; critic gradients are optimized separately from the joint actor and personalized-embedding gradients.

\begin{algorithm*}[t]
\caption{``Interact-Collect-Optimize'' Loop of COPE}
\label{alg:optimization}
\begin{algorithmic}[1]
\REQUIRE User set $U$, Initial Model Parameters $\theta_1$, Number of Sessions $T$, Feedback Probability $p$
\STATE Initialize personalized embeddings $\mathcal{V}_1$ randomly
\STATE Initialize critics $V_{\phi_{\mathrm{eval}}}$ and $V_{\phi_{\mathrm{resp}}}$ from the assistant model
\FOR{session $t = 1$ to $T$}
    \STATE \textbf{// Interaction \& Collection}
    \STATE Collect queries $\{q_i^t\}_{i=1}^N$
    \STATE Generate responses $a_i^t \sim \pi_{\theta_t}(\cdot | \mathbf{v}_i^t, q_i^t)$ for all users
    \STATE Receive sparse feedback: $f_i^t, s_i^t$ with probability $p$ for all users
    \STATE Construct datasets $\mathcal{D}_1^t$ ($i\in{\Omega_1^t}$, no feedback) and $\mathcal{D}_2^t$ ($i\in{\Omega_2^t}$, with feedback), such that $|\Omega_1^t| + |\Omega_2^t| = N$
    
    \STATE \textbf{// Unified Optimization Step}
    \STATE \textit{\textbf{Step 1: Compute Gradients for Preference Capture (SFT)}}
    \STATE \quad Calculate $\nabla_{\mathbf{v}} \mathcal{L}_{\text{SFT}}$ via Eq. \ref{eq:sft_loss} (for $i\in\Omega_2^t$)
    
    \STATE \textit{\textbf{Step 2: Compute Gradients for Self-Evaluation Calibration (RL)}}
    \STATE \quad Generate self-reflection and score $\hat{s}_i^t$ using $\pi_{\theta_t}$ and $\mathbf{v}_i^t$ for $i\in\Omega_1^t\cup\Omega_2^t$
    \STATE \quad Compute reward $R_{\text{eval}}(\hat{s}_i^t, s_i^t)$ for $i\in\Omega_2^t$
    \STATE \quad Compute returns and advantages using $V_{\phi_{\mathrm{eval}}}$; update $\phi_{\mathrm{eval}}$ via Eq.~\ref{eq:critic_loss}
    \STATE \quad Calculate $\nabla_{\theta, \mathbf{v}} \mathcal{J}_{\text{eval}}$ via PPO (for $i\in\Omega_2^t$)
    
    \STATE \textit{\textbf{Step 3: Compute Gradients for Personalized Response Optimization (RL)}}
    \STATE \quad Evaluate completeness $r_{i,\text{comp}}^t$ and personalization $r_{i,\text{pers}}^t$ for $i\in\Omega_1^t\cup\Omega_2^t$
    \STATE \quad Compute final reward $r_i^t = r_{i,\text{comp}}^t \cdot r_{i,\text{pers}}^t$ for $i\in\Omega_1^t\cup\Omega_2^t$
    \STATE \quad Compute returns and advantages using $V_{\phi_{\mathrm{resp}}}$; update $\phi_{\mathrm{resp}}$ via Eq.~\ref{eq:critic_loss}
    \STATE \quad Calculate $\nabla_{\theta} \mathcal{J}_{\text{policy}}$ via PPO (for $i\in\Omega_1^t\cup\Omega_2^t$)
    
    \STATE \textit{\textbf{Step 4: Joint Parameter Update}}
    \STATE \quad Update $\theta_t \to \theta_{t+1}$ using aggregated gradients $\nabla_{\theta} \mathcal{J}_{\text{eval}} + \nabla_{\theta} \mathcal{J}_{\text{policy}}$
    \STATE \quad Update $\mathbf{v}_i^t \to \mathbf{v}_i^{t+1}$ using aggregated gradients $\nabla_{\mathbf{v}} \mathcal{L}_{\text{SFT}} + \nabla_{\mathbf{v}} \mathcal{J}_{\text{eval}}$
\ENDFOR
\end{algorithmic}
\end{algorithm*}

\section{Extension to Multi-Turn Sessions}
\label{app:multi_turn_extension}
{
Our experiments use single-turn sessions because longitudinal multi-turn personalization data are difficult to obtain under user-privacy constraints. Nevertheless, COPE can be extended to multi-turn interactions with session-level feedback without changing its three-objective design. For user $u_i$ in session $t$, we replace the single-turn tuple $(q_i^t,a_i^t)$ with a dialogue trajectory
\begin{equation}
    h_i^t = \{(q_{i,k}^t,a_{i,k}^t)\}_{k=1}^{K},
\end{equation}
where $K$ is the number of turns in the session. At turn $k$, the assistant generates
\begin{equation}
    a_{i,k}^t \sim \pi_{\theta_t}
    \left(\cdot \mid \mathbf{v}_i^t, h_{i,<k}^t, q_{i,k}^t\right),
\end{equation}
where $h_{i,<k}^t$ is the preceding dialogue history. The same personalized embedding $\mathbf{v}_i^t$ is used across all turns to represent preferences accumulated across sessions, while $h_{i,<k}^t$ provides the local context within the current session. After the final turn, the user provides one textual feedback message $f_i^t$ and one overall satisfaction score $s_i^t$ with probability $p$. The collected datasets become $\mathcal{D}_1^t=\{h_i^t\}_{i\in\Omega_1^t}$ for sessions without feedback and $\mathcal{D}_2^t=\{(h_i^t,f_i^t,s_i^t)\}_{i\in\Omega_2^t}$ for sessions with feedback.

For preference capture, the complete dialogue replaces the single query--response pair as the context for predicting the session-level textual feedback:
\begin{equation}
\begin{aligned}
    \mathcal{L}_{\mathrm{SFT}}^{\mathrm{multi}}
    ={}& -\mathbb{E}_{i\in\Omega_2^t}\Bigg[
    \sum_{j=1}^{|f_i^t|}\log \pi_{\theta_t}\Big(
    f_{i,j}^t \mid {}\\
    &\hspace{32mm}\mathbf{v}_i^t,h_i^t,f_{i,<j}^t\Big)\Bigg].
\end{aligned}
\end{equation}
As in the single-turn formulation, this objective updates only the personalized embedding. For self-evaluation calibration, the model predicts an overall score $\hat{s}_i^t$ from $(\mathbf{v}_i^t,h_i^t)$, optionally conditioned on $f_i^t$ under the same teacher-forcing strategy, and is optimized against $s_i^t$ using the reward in Eq.~\ref{eq:r_eval}. Thus, the self-evaluator learns to assess the personalization quality of the complete session rather than a single response.

For personalized response optimization, an evaluator first assigns a session-level completeness reward $r_{i,\mathrm{comp}}^t$. The personalization reward is the real session score when feedback is available and the predicted session score otherwise:
\begin{equation}
\begin{aligned}
    r_{i,\mathrm{pers}}^t =
    \begin{cases}
        \hat{s}_i^t, & i\in\Omega_1^t,\\
        s_i^t, & i\in\Omega_2^t.
    \end{cases}\\
    r_{i,\mathrm{sess}}^t
    = r_{i,\mathrm{comp}}^t \cdot r_{i,\mathrm{pers}}^t.
\end{aligned}
\end{equation}
The complete dialogue is treated as one trajectory, with $r_{i,\mathrm{sess}}^t$ serving as its terminal reward. For brevity, let $c_{i,k}^t=(\mathbf{v}_i^t,h_{i,<k}^t,q_{i,k}^t)$ denote the personalized context at turn $k$, and define
\begin{equation}
    d_{i,k}^t = \mathrm{KL}\!\left(
    \pi_{\theta_t}(\cdot\mid c_{i,k}^t)
    \|\pi_{\mathrm{ref}}(\cdot\mid c_{i,k}^t)\right).
\end{equation}
The corresponding policy objective is
\begin{equation}
    \mathcal{J}_{\mathrm{policy}}^{\mathrm{multi}}
    = \mathbb{E}_{i\in\Omega_1^t\cup\Omega_2^t}
    \left[r_{i,\mathrm{sess}}^t
    - \beta \sum_{k=1}^{K} d_{i,k}^t\right].
\end{equation}
The three objectives can then be aggregated in the same unified update described in Section~\ref{sec:methodology}. Therefore, the main change is the optimization unit: a complete dialogue session replaces a single interaction, while COPE's personalized embedding, sparse-feedback mechanism, and joint optimization remain unchanged.
}

\begin{figure*}[ht]
    \centering
    \includegraphics[width=\linewidth]{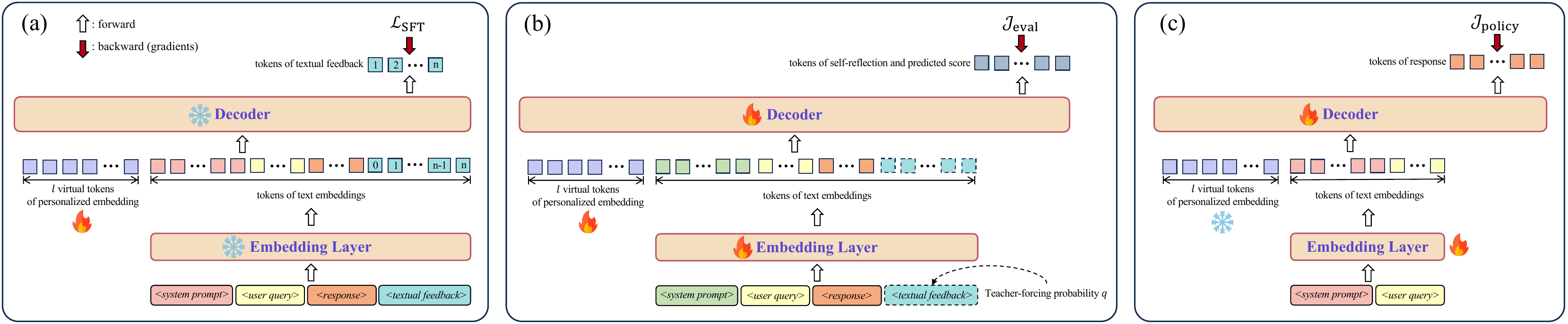}
    \caption{Schematic illustration of personalized embedding integration and the forward computation and backward optimization of the three objectives in the optimization stage of COPE.
    Personalized embeddings are implemented as $l$ virtual tokens and prepended to the text embeddings to provide user-specific guidance.
    For clarity, the assistant model is conceptually decomposed into an embedding layer and a decoder to explicitly visualize the embedding of textual inputs and their concatenation with personalized embeddings, although both components belong to the same assistant model.
    Two different system prompts are used: a response-generation prompt (pink) and a self-evaluation prompt (green), both of which are provided in our code repository.
    Trainable parameters under each objective are marked with spark icons, while frozen parameters are indicated by snowflake icons.}
    \label{fig:pvec}
\end{figure*}

\section{Illustration of Personalized Embedding Integration and Optimization Flow}
\label{app:pvec}
Figure~\ref{fig:pvec} illustrates the optimization stage of COPE, showing how personalized embeddings are prepended to text embeddings and how the three objectives are computed and optimized.

\section{Chronological Traversal Protocol}
\label{app:traversal_protocol}
\begin{figure*}[ht]
    \centering
    \includegraphics[width=\linewidth]{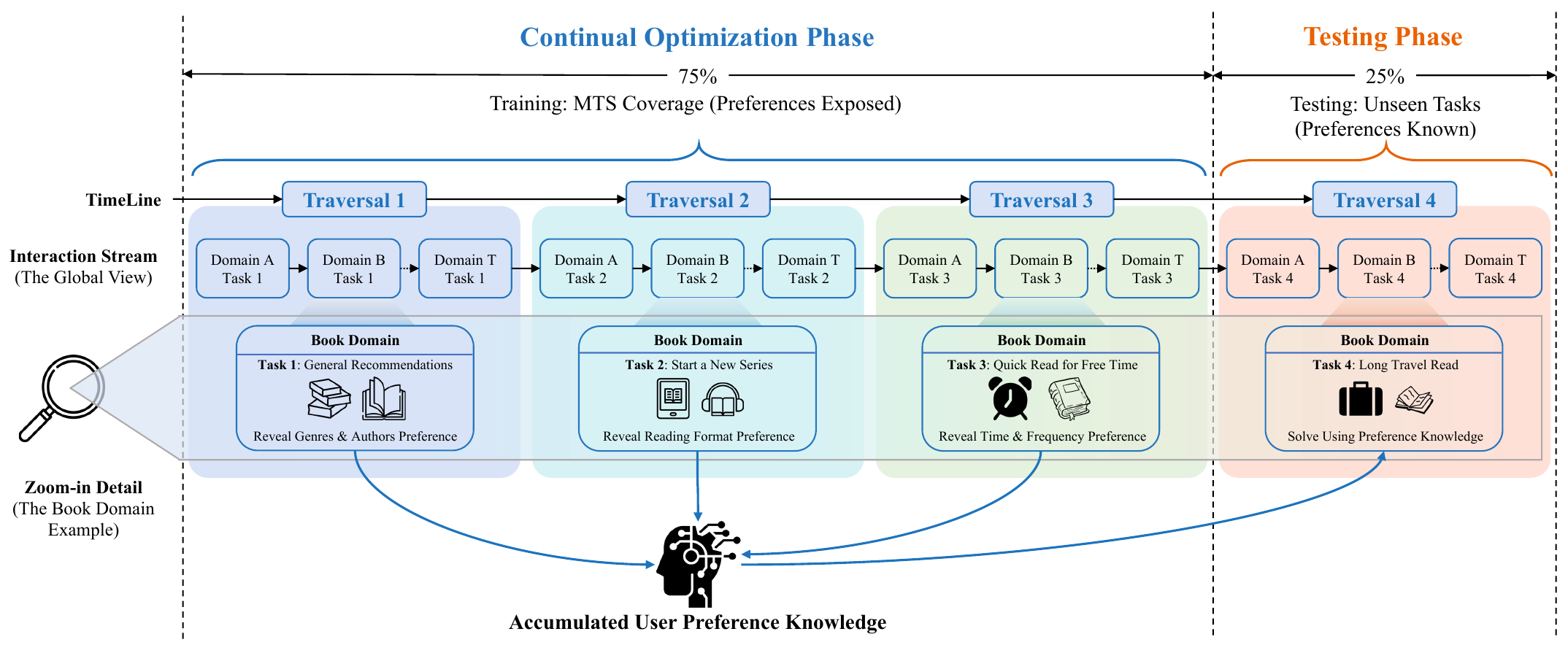}
    \caption{Chronological traversal and preference coverage. The experiment proceeds through four traversals: Traversals $1$-$3$ continually expose user preferences via MTS tasks across $17$ domains, while Traversal $4$ evaluates generalization on a hold-out task. The illustration shows one user for clarity; all users follow the same protocol in parallel.}
    \label{fig:traversal_protocol}
\end{figure*}

Figure~\ref{fig:traversal_protocol} illustrates the chronological traversal process used in our experimental setup.

\section{Decomposed Main Results}
\label{app:decomposed_main_results}
Table~\ref{tab:decomposed_main_results} reports the completeness and personalization components underlying the product scores in Table~\ref{tab:main_results}.

\begin{table}[htbp]
    \centering
    \caption{Decomposed completeness and personalization scores under varying user feedback probabilities ($\text{mean}\pm\text{std}$). $\dagger$ indicates methods that involve training.}
    \label{tab:decomposed_main_results}
    \small
    \setlength{\tabcolsep}{2pt}
    \renewcommand{\arraystretch}{1.16}
    \begin{tabular}{@{}l c c c@{}}
        \toprule
        \multirow{2}{*}{\textbf{Methods}} & \multicolumn{3}{c}{\textbf{User Feedback Probability}} \\
        \cmidrule(lr){2-4}
         & $p=0.25$ & $p=0.50$ & $p=0.75$ \\ \hline
        \multicolumn{4}{c}{\textbf{Completeness}} \\ \hline
        Base & $0.78{\pm}0.00$ & $0.78{\pm}0.00$ & $0.78{\pm}0.00$ \\ \hline
        SR & $0.87{\pm}0.01$ & $0.87{\pm}0.00$ & $0.87{\pm}0.00$ \\ \hline
        DR & $0.85{\pm}0.00$ & $0.86{\pm}0.01$ & $0.86{\pm}0.00$ \\ \hline
        Q-DR & $0.85{\pm}0.01$ & $0.85{\pm}0.00$ & $0.85{\pm}0.01$ \\ \hline
        PAP & $0.90{\pm}0.01$ & $0.91{\pm}0.00$ & $0.91{\pm}0.00$ \\ \hline
        T-PAP$^\dagger$ & $0.95{\pm}0.00$ & $0.95{\pm}0.00$ & $0.96{\pm}0.00$ \\ \hline
        COPE$^\dagger$ & $\textbf{0.98}{\pm}0.01$ & $\textbf{0.99}{\pm}0.00$ & $\textbf{0.99}{\pm}0.00$ \\ \hline
        COPE$^\dagger$+SR & $0.97{\pm}0.00$ & $0.98{\pm}0.00$ & $0.98{\pm}0.00$ \\ \hline
        \multicolumn{4}{c}{\textbf{Personalization}} \\ \hline
        Base & $2.69{\pm}0.01$ & $2.69{\pm}0.01$ & $2.69{\pm}0.01$ \\ \hline
        SR & $3.35{\pm}0.04$ & $3.37{\pm}0.01$ & $3.40{\pm}0.03$ \\ \hline
        DR & $3.36{\pm}0.03$ & $3.44{\pm}0.02$ & $3.55{\pm}0.02$ \\ \hline
        Q-DR & $3.51{\pm}0.06$ & $3.70{\pm}0.04$ & $3.82{\pm}0.04$ \\ \hline
        PAP & $3.85{\pm}0.01$ & $3.93{\pm}0.02$ & $3.96{\pm}0.00$ \\ \hline
        T-PAP$^\dagger$ & $4.59{\pm}0.03$ & $4.62{\pm}0.03$ & $4.83{\pm}0.01$ \\ \hline
        COPE$^\dagger$ & $4.62{\pm}0.06$ & $4.73{\pm}0.05$ & $4.78{\pm}0.02$ \\ \hline
        COPE$^\dagger$+SR & $\textbf{5.20}{\pm}0.02$ & $\textbf{5.31}{\pm}0.05$ & $\textbf{5.43}{\pm}0.03$ \\
        \bottomrule
    \end{tabular}
\end{table}

The decomposed results show that optimization improves both task completion and preference alignment. Compared with Base, all personalization methods achieve higher completeness and personalization scores, and training-based methods further strengthen both components. COPE obtains the highest completeness across all feedback probabilities, while COPE+SR achieves the strongest personalization scores and remains close to COPE in completeness. These trends indicate that learned user embeddings and retrieved histories provide complementary signals for improving both dimensions of personalized response quality.

\section{Implementation Details}
\label{app:implementation_details}
\subsection{COPE Implementation Details}
\label{app:cope_implementation_details}
We use Qwen3-1.7B~\citet{yang2025qwen3} as the assistant model and Qwen-Flash~\citet{alibaba2026qwenflash} as the user simulator in the main experiments. We pre-generate the initial query and rubrics for each user-task instance from the user's true preferences $\eta_i$; fixing these artifacts before interaction reduces simulator randomness and ensures that both user behavior and feedback criteria consistently reflect the hidden preferences. We manually inspect the pre-generated query-rubric pairs to check that they reflect the underlying user preferences. Thus, during interaction, the user simulator poses the pre-generated query and, with probability $p$, provides evaluative feedback on the assistant's response. Appendix~\ref{app:simulator_validity} further discusses the rationale for this simulator-based protocol.
During evaluation, we use the pre-generated rubrics to assess the assistant's response. Each rubric item specifies a condition the response should satisfy and an associated score. Since the rubrics are fixed in advance and manually inspected for consistency with the user's preferences, the user model only needs to judge the response against these rubrics, enabling more stable feedback that remains aligned with the user's preferences. The simulator evaluates the response against these rubrics to generate textual feedback $f_i^t$, summarizing satisfied or violated criteria, and a numerical score $s_i^t$, computed as the sum of scores for all satisfied items. A concrete example of the interaction, feedback, and rubrics, is provided in Appendix~\ref{app:interaction_case}.
To reflect the coarse granularity of real-world feedback, we compress the raw integer feedback ($0$-$10$) into a coarser four-point scale ($0$-$3$). Specifically, original scores $\{0, 1, 2\}$ are mapped to $0$, $\{3, 4, 5\}$ to $1$, $\{6, 7, 8\}$ to $2$, and $\{9, 10\}$ to $3$. Consequently, the self-evaluation mechanism predicts these mapped scores to compute matching rewards, and the mapped scores subsequently serve as the reward signal for personalized response optimization. However, when evaluating model performance, we still report results using the original (pre-mapping) scores.
Regarding hyperparameters, we set the learning rate to $1\times10^{-3}$ for personalized embeddings and $1\times10^{-6}$ for model parameters, with a KL penalty weight $\beta$ of $0.01$. Each personalized embedding consists of $10$ learnable virtual tokens (i.e., $l=10$) and is randomly initialized prior to the training process. The two critics are independently initialized from the assistant model's weights and warmed up for $5$ steps. The teacher-forcing probability $q$ during self-evaluation calibration is set to $0.25$.
Our algorithm is implemented based on the VeRL framework, with corresponding adaptations to enable efficient inference using SGLang with personalized embedding prefixes. Our implementation, associated datasets, and all prompts used in this study are available at \url{https://github.com/Quark-Medical/COPE}.

\subsection{Rationale for Simulator-Based Evaluation}
\label{app:simulator_validity}
Although our experimental setting is motivated by real user--assistant interactions, recruiting a large number of users to participate in the full continual optimization loop would be costly and would require careful privacy protection, especially when persistent user representations are updated over $51$ chronological interactions. For reproducible and efficient algorithm development, we therefore follow the common practice of using LLM-based user simulation. Specifically, we instantiate users from PersonaLens~\citep{zhao-etal-2025-personalens}, a public benchmark designed for personalization evaluation in task-oriented conversational assistants, with diverse profiles, domain-specific preferences, interaction histories, situational contexts, and validation of profile consistency, preference distributions, lexical diversity, and user-agent quality.

Our simulator is also constrained by pre-generated and manually inspected query-rubric pairs. During interaction, the user simulator only presents the fixed query and judges the response against the fixed rubric when feedback is sampled, rather than freely inventing new preferences or criteria. This makes feedback generation a more stable instruction-following and rubric-checking task, while keeping the evaluation criteria aligned with the hidden profile and comparable across methods. Simulator-based user interaction is also common in recent personalized alignment work such as RLPA~\citep{zhao2025teaching}. We therefore view this setting as a controlled benchmark-derived approximation of sparse personalized interaction, while leaving real-user validation to future privacy-preserving studies.

\subsection{Artifacts, Licenses, and Compute}
\label{app:artifacts_licenses_compute}
We use PersonaLens as a public research benchmark for evaluating personalization, rather than as deployed user data. PersonaLens is released under the CC-BY-NC-4.0 license, contains English-language task-oriented conversational data, and provides rich user profiles, interaction histories, and $20$ domains with $4$ to $5$ tasks per domain. We use Qwen3-1.7B under its Apache-2.0 license. Qwen-Flash and DeepSeek-V3 are accessed through their respective API services for simulation and robustness evaluation. All artifacts are used for research and evaluation purposes in accordance with their documented access conditions. The public repository is intended for research-use reproducibility and will follow release terms compatible with the upstream non-commercial artifacts.

We do not collect new real-user interaction data or recruit human participants. Our experiments use benchmark-derived user profiles and simulator-generated interactions. We manually inspect the generated query-rubric pairs for consistency with the underlying preferences and for obvious personally identifying information and unsafe or offensive content. No additional anonymization is applied beyond using benchmark-provided user identifiers and simulator-generated interaction artifacts.

Our implementation uses VeRL 0.6.0, SGLang 0.4.9, \texttt{rank-bm25} for BM25 retrieval, and \texttt{lm-eval} v0.4.3 for general capability evaluation.

\subsection{Comparison Method Implementation Details}
\label{app:comparison_method_implementation_details}
For \textbf{PAP}, a separate model (Qwen-Flash~\citet{alibaba2026qwenflash}) first summarizes the user's interaction history from the first three domain traversals to build a user profile; the profile is then concatenated with the prompt to guide the assistant model in generating the response.
\textbf{RAP} enhances personalization by retrieving historical interactions relevant to the user's current query. In our experiments, we retrieve the top-$6$ historical interactions. \textbf{SR} uses BM25~\citet{trotman2014improvements} for sparse retrieval, while \textbf{DR} uses Qwen3-Embedding~\citet{zhang2025qwen3} to encode both the interaction history and the user's query and then performs retrieval based on embedding similarity. \textbf{Q-DR} is a query-expansion variant of dense retrieval, where the assistant model first generates an auxiliary intermediate response conditioned on the initial query; this intermediate response is then embedded and used to retrieve the top-$6$ relevant historical interactions.
For \textbf{T-PAP}, we follow PAP by incorporating the user profile summarized by Qwen-Flash~\citet{alibaba2026qwenflash} as part of the prompt. To ensure a fair comparison with COPE, T-PAP employs the same optimization objectives as COPE, namely Self-Evaluation Calibration (RL) and Personalized Response Optimization (RL), both guided by the summarized user profiles. Since T-PAP does not include learnable personalized embeddings, it omits the Preference Capture (SFT) objective.
\textbf{COPE+SR} combines the COPE-optimized model with the same SR module described above, allowing the model to use both learned personalized embeddings and retrieved historical interactions during response generation.

\section{Interaction Case}\label{app:interaction_case}
\begin{figure}[ht]
    \centering
    \includegraphics[width=0.9\linewidth]{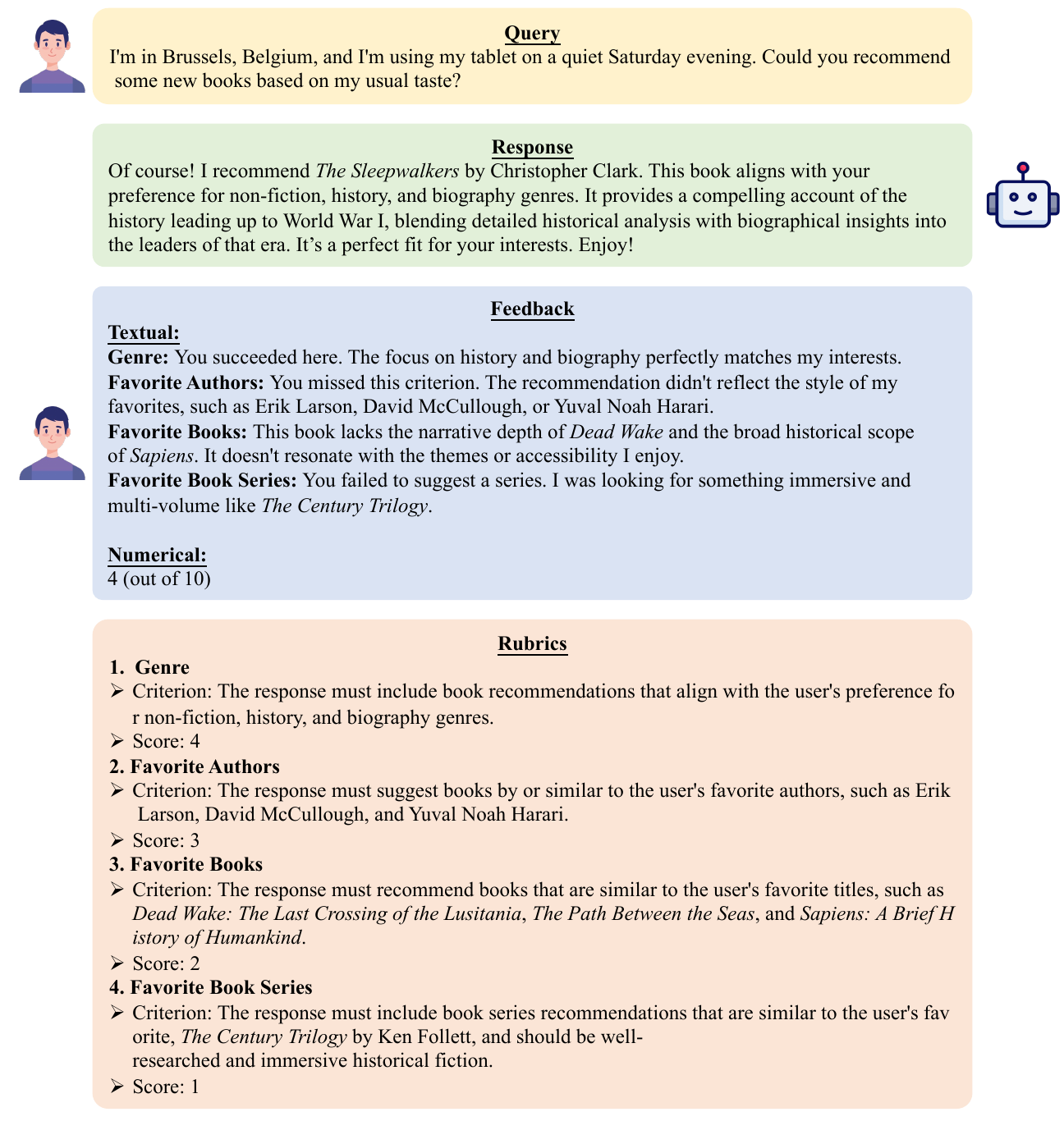}
    \caption{An illustrative example of the interaction between the user simulator and the assistant. The figure shows the pre-generated user query, the assistant's response, the rubrics, and the user feedback. The feedback includes textual feedback summarizing satisfied or violated criteria, and a numerical feedback computed as the sum of scores for all satisfied items.}
    \label{fig:case_study}
\end{figure}

In this section, we provide an example comprising a single interaction turn between the user simulator (Qwen-Flash~\citet{alibaba2026qwenflash}) and the assistant (Qwen3-1.7B~\citet{yang2025qwen3}), along with the corresponding pre-generated initial query and rubrics and the user-generated feedback, in Figure~\ref{fig:case_study}.

\section{Instance-Level Self-Evaluation Reliability}
\label{app:instance_level_self_eval}
The Pearson correlation reported in Section~\ref{sec:experiments} is computed from step-level aggregates and could in principle hide instance-level disagreements. To examine reliability at the level of individual interactions, we conduct an additional pairwise ranking analysis. Specifically, we randomly select a subset of the held-out validation interactions and run the assistant on each twice, yielding $1{,}133$ pairs of responses where the user provides real feedback on both runs. Both sides are compared in the same $0$--$3$ grade space: the user's raw $0$--$10$ score is mapped via the same discretization the assistant is trained against.

We discard pairs where the user assigns the same grade to both runs, since these pairs have no ground-truth ordering. The remaining user-ranked pairs are classified into three categories:
\begin{itemize}
    \item \textbf{Concordant}: the proxy ranks the two runs in the same direction as the user.
    \item \textbf{Discordant}: the proxy ranks them in the opposite direction.
    \item \textbf{Proxy silent}: the proxy assigns the same grade to both runs (the user can rank, the proxy cannot).
\end{itemize}

Among the $1{,}133$ pairs, $710$ are discarded as user-tied, leaving $423$ user-ranked pairs ($207$ concordant, $78$ discordant, $138$ proxy-silent). The proxy gives a directional judgment on $67.4\%$ of these pairs; the remaining $32.6\%$ are proxy-silent and reflect the inherent granularity loss of the four-grade discretization. Among the $285$ pairs where the proxy does produce a ranking, its direction agrees with the user's $72.63\%$ of the time. This indicates that, in the cases where the proxy is decisive, it carries non-trivial directional signal at the instance level.

\subsection{Why Self-Evaluation Calibration Works}
\label{app:self_eval_mechanism}
Self-evaluation calibration combines preference representation and score calibration. Preference Capture SFT encodes textual feedback into each personalized embedding, teacher forcing grounds response properties in user judgments, and PPO jointly updates the embedding and LLM against numerical feedback. Thus, the embedding represents what the user values, while the LLM maps response quality under those preferences to a user-specific score. Supporting this interpretation, adding $\mathcal{J}_{\mathrm{eval}}$ and then $\mathcal{L}_{\mathrm{SFT}}$ improves performance from $3.36$ to $4.57$ and $4.69$ (Table~\ref{tab:objective_ablation}); within $\mathcal{J}_{\mathrm{eval}}$, updating only the LLM or embedding yields $4.49$ and $4.53$, versus $4.69$ when both are updated (Table~\ref{tab:parameters_ablation}).

To complement the aggregate results, we analyze two held-out self-evaluation outputs for which the current textual feedback is unavailable. In one shopping interaction, the self-evaluator recognizes that the response covers the user's preferred product categories and attributes such as recycled materials, ethical production, and cruelty-free products, and assigns the same highest coarse grade as the mapped user feedback. In a merchandise interaction, it detects that the response recommends San Lorenzo products despite the user's established preference for Club Universidad de Chile, and assigns the same lowest grade as the mapped feedback. These examples illustrate how the self-evaluator identifies both preference matches and conflicts, showing how training-time feedback and accumulated user preferences can support self-evaluation on unseen interactions.

\section{General Capability Evaluation}
\label{app:general_capability}
To evaluate whether continual personalization causes catastrophic forgetting of general capabilities, we assess the optimized model after the $51$-interaction optimization phase described in Section~\ref{sec:experiments}. We use \texttt{lm-eval}~\citet{eval-harness} to evaluate instruction following with IFEval~\citet{zhou2023instructionfollowingevaluationlargelanguage}, mathematical reasoning with GSM8K~\citet{cobbe2021gsm8k}, and multi-domain knowledge understanding with MMLU~\citet{hendryckstest2021, hendrycks2021ethics}. Table~\ref{tab:general_capability} compares the optimized model with the original Qwen3-1.7B~\citet{yang2025qwen3}.

\begin{table*}[t]
    \centering
    \caption{General capability evaluation before and after continual personalization.}
    \label{tab:general_capability}
    \begin{tabular}{@{}l l c c@{}}
        \toprule
        Benchmark & Metric & Qwen3-1.7B & Optimized model \\
        \midrule
        \multirow{4}{*}{IFEval}
        & inst\_level\_loose\_acc & $73.50$ & $74.58$ \\
        & inst\_level\_strict\_acc & $70.50$ & $71.10$ \\
        & prompt\_level\_loose\_acc & $65.43{\pm}2.05$ & $67.47{\pm}2.02$ \\
        & prompt\_level\_strict\_acc & $61.74{\pm}2.09$ & $62.66{\pm}2.08$ \\
        \midrule
        \multirow{2}{*}{GSM8K}
        & flexible-extract & $68.69{\pm}1.28$ & $69.37{\pm}1.27$ \\
        & strict-match & $68.31{\pm}1.28$ & $69.14{\pm}1.27$ \\
        \midrule
        \multirow{5}{*}{MMLU}
        & humanities & $52.73{\pm}0.70$ & $52.50{\pm}0.70$ \\
        & other & $64.15{\pm}0.84$ & $64.11{\pm}0.84$ \\
        & social sciences & $68.28{\pm}0.82$ & $67.99{\pm}0.83$ \\
        & stem & $59.21{\pm}0.85$ & $59.02{\pm}0.85$ \\
        & overall & $60.12{\pm}0.40$ & $59.93{\pm}0.40$ \\
        \bottomrule
    \end{tabular}
\end{table*}

The optimized model shows small numerical increases on IFEval and GSM8K, while the MMLU overall score decreases only marginally from $60.12$ to $59.93$. However, these changes are all within one standard deviation of the corresponding benchmark results, so we interpret them as comparable performance rather than evidence of broad general-capability improvement or degradation. Overall, COPE's continual personalization improves personalized behavior while maintaining general instruction-following, reasoning, and knowledge understanding capabilities in this setting, suggesting that it does not induce catastrophic forgetting over the evaluated training horizon.

\section{Preference Shift Adaptation Evaluation}
\label{app:preference_shift}
Real-world users may change their preferences over time. We therefore evaluate COPE under two simulated preference-shift settings. Both settings use the same $256$ users as in the main experiments: the within-domain shift splits these users into two groups while keeping the domain coverage fixed, and the new-domain shift samples its two $64$-user groups from non-overlapping domain sets. In both settings, the model continues training after the simulated shift, and the personalized embedding associated with each user slot is carried over and further updated rather than re-initialized. This tests whether the existing user representation can be revised when the underlying preference distribution changes. For the curves, checkpoints before the shift are evaluated on the validation tasks of the pre-shift users, while checkpoints after the shift are evaluated on the validation tasks of the post-shift users. Each checkpoint is evaluated three times, and the plotted point and error bar report the mean and standard deviation across these validation runs.

In the within-domain shift setting, we randomly split the users into two groups of $128$ users. During the first phase, COPE is trained for $51$ interactions using tasks from the first user group. After this point, each user slot is switched to a user from the second group to simulate a shift in the underlying preference distribution, and training continues with the same personalized embedding. As shown in Figure~\ref{fig:user_preference_shift}, the personalization score decreases slightly after the shift, but quickly recovers and continues to improve with further training. In contrast, the completeness score is largely unaffected by the shift and remains at a high level throughout the second phase. These results indicate that COPE can revise the carried-over personalized embeddings without losing task completeness when preferences shift within the seen domain coverage.

In the new-domain shift setting, we further stress-test adaptation to new preference domains. We first randomly choose $8$ domains and sample $64$ users interested in these domains. We then choose another $8$ domains that do not overlap with the first set and sample another $64$ users interested in the new domains. COPE is trained for $24$ interactions on the first group and then continues training on the second group using the carried-over personalized embedding for each user slot. As shown in Figure~\ref{fig:domain_preference_shift}, both the personalization and completeness scores decrease after the shift, which is expected because the model is exposed to a new set of domains. As training proceeds, both scores consistently increase, showing that COPE can adapt not only to preference changes within the same broad environment, but also to shifts involving substantially different domains. The recovery after the shift further suggests that COPE can adapt to new preferences even when the relevant domains change substantially.

\section{Cross-Family User Model Evaluation}
\label{app:user_model_robustness}
To test whether the observed gains are tied to the Qwen-family user simulator, we conduct an additional evaluation by replacing the user model with DeepSeek-V3~\citet{liu2024deepseek}. The assistant model, chronological traversal protocol, and comparison methods remain unchanged. We use the setting $p=0.5$ and $q=0.5$, where $p$ is the probability of receiving user feedback and $q$ is the teacher-forcing probability in self-evaluation calibration. Initial queries and rubrics are still pre-generated from user preferences, while DeepSeek-V3 provides user feedback during interaction and evaluates both personalization and task completeness.

\begin{table}[htbp]
    \centering
    \caption{Cross-family user model evaluation with DeepSeek-V3 as the user simulator ($p=0.5$, $q=0.5$). Scores are personalization score $\times$ binary completeness score, out of $10$ ($\text{mean}\pm\text{std}$). $\dagger$ indicates methods that involve training.}
    \label{tab:deepseek_user_model}
    \begin{tabular}{l c}
        \toprule
        Methods & Scores \\
        \midrule
        Base & $2.16{\pm}0.02$ \\
        SR & $3.00{\pm}0.01$ \\
        DR & $3.08{\pm}0.03$ \\
        Q-DR & $3.30{\pm}0.01$ \\
        PAP & $3.31{\pm}0.03$ \\
        T-PAP$^\dagger$ & $4.06{\pm}0.02$ \\
        COPE$^\dagger$ & $\textbf{4.23}{\pm}0.06$ \\
        \bottomrule
    \end{tabular}
\end{table}

Compared with the Qwen-Flash~\citet{alibaba2026qwenflash} user setting, DeepSeek-V3 yields lower absolute scores across methods, which is attributable to its stricter criteria when judging task completeness and preference alignment. Nevertheless, the overall ordering is preserved: retrieval and profile-based prompting improve over the base assistant, training-based personalization further improves performance, and COPE obtains the highest score. These results indicate that COPE remains effective under different user-model judging styles.

\section{Independent Blinded Pairwise Evaluation}
\label{app:third_party_judge}
To address the concern that COPE's gains may reflect fitting the Qwen-family simulator and its rubric-based scoring rather than genuine preference adaptation, we conduct an independent, blinded pairwise evaluation of COPE and T-PAP responses ($p=0.5$) using a human annotator. We randomly sample $12$ aligned COPE/T-PAP response pairs from each of the $17$ held-out validation steps in the fourth traversal, yielding $204$ pairs in total. For each pair, we randomize the presentation order of the two responses to mitigate position bias and ask the human annotator to select the response that better matches the user. The annotator sees only the user's demographics, the affinities relevant to the queried domain, the query, and the two responses; the simulator's rubrics and numerical scores are withheld.

\begin{figure}[h]
    \centering
    \includegraphics[width=\linewidth]{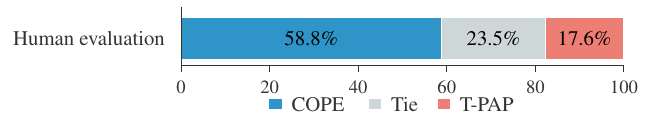}
    \caption{Blinded pairwise preferences between COPE and T-PAP from human evaluation. Percentages are computed over all $204$ randomly sampled validation interactions.}
    \label{fig:third_party_judge}
\end{figure}

Across the $204$ pairs, the human evaluation prefers COPE on $120$ pairs and T-PAP on $36$, with $48$ ties. Thus, COPE wins $76.92\%$ of the decisive comparisons. The consistent preference for COPE in the blinded, rubric-free human evaluation demonstrates that COPE adapts to human preferences rather than merely learning the simulator's scoring behavior.


\end{document}